\pdfoutput=1
\documentclass[11pt]{article}
\usepackage{acl}
\usepackage{times}
\usepackage{latexsym}
\usepackage[T1]{fontenc}
\usepackage[utf8]{inputenc}
\usepackage{microtype}
\usepackage{inconsolata}
\usepackage[utf8]{inputenc}
\usepackage{microtype}
\usepackage{graphicx}
\usepackage{subfigure}
\usepackage{booktabs}
\usepackage{hyperref}
\usepackage[algo2e,ruled,vlined]{algorithm2e}
\usepackage{algorithm}
\usepackage{algorithmic}

\usepackage{pifont}
\usepackage{float}
\newfloat{eqn}{htbp}{loe}
\floatname{eqn}{Equation}
\usepackage{tabularx}
\usepackage{amssymb}
\usepackage{breqn}
\usepackage{amsthm}
\usepackage[nopar]{lipsum}
\usepackage{changepage}
\usepackage{multirow}
\theoremstyle{definition}
\newtheorem{definition}{Definition}
\usepackage{multicol}
\makeatletter
\newcommand\fs@nobottomruled{\def\@fs@cfont{\bfseries}\let\@fs@capt\floatc@ruled
  \def\@fs@pre{\hrule height.8pt depth0pt \kern2pt}%
  \def\@fs@post{}
  \def\@fs@mid{\kern2pt\hrule\kern2pt}%
  \let\@fs@iftopcapt\iftrue}
\makeatother
\floatstyle{nobottomruled}
\restylefloat{algorithm}

\definecolor{cadmiumgreen}{rgb}{0.0, 0.42, 0.24}

\DeclareMathOperator*{\argmax}{arg\,max}

\usepackage{centernot}

\usepackage{collectbox}
\makeatletter
\newcommand{\mybox}{%
    \collectbox{%
        \setlength{\fboxsep}{1pt}%
        \fbox{\BOXCONTENT}%
    }%
}

\usepackage{nicefrac}       
\usepackage{thm-restate}
\usepackage{amsmath,amsfonts,bm}
\usepackage{amssymb,amsmath}
\usepackage{amsthm}
\usepackage[scr=boondoxo]{mathalfa}

\usepackage{listings}

\definecolor{codegreen}{rgb}{0.36, 0.54, 0.66}
\definecolor{codegray}{rgb}{1,1,1}
\definecolor{codepurple}{rgb}{0.58,0,0.82}
\definecolor{bluepigment}{rgb}{0.2, 0.2, 0.6}
\definecolor{amethyst}{rgb}{0.6, 0.4, 0.8}
\definecolor{backcolour}{rgb}{1,1,1}
\lstdefinestyle{mystyle}{
    backgroundcolor=\color{backcolour},   
    numberstyle=\tiny\color{codegray},
    basicstyle=\ttfamily\footnotesize,
    breakatwhitespace=false,         
    breaklines=true,                 
    captionpos=b,                    
    keepspaces=true,                 
    numbers=left,                    
    numbersep=5pt,                  
    showspaces=false,                
    showstringspaces=false,
    showtabs=false,                  
    tabsize=2
}
\usepackage{cases}
\usepackage{enumitem}
\usepackage[scaled=.8]{beramono}
\usepackage{url}
\usepackage{tikz}
\newcommand*\circled[1]{\tikz[baseline=(char.base)]{
            \node[shape=circle,draw,inner sep=0.5pt] (char) {#1};}}

\usepackage[symbol]{footmisc}

\title{\textit{Learn2Weight}: Parameter Adaptation against \\ Similar-domain Adversarial Attacks}

\author{Siddhartha Datta \\
  University of Oxford \\ 
  \texttt{siddhartha.datta@cs.ox.ac.uk} \\
}

\begin{document}
\maketitle
\begin{abstract}
Recent work in black-box adversarial attacks for NLP systems has attracted much attention. Prior black-box attacks assume that attackers can observe output labels from target models based on selected inputs. In this work, inspired by  adversarial transferability, we propose a new type of black-box NLP adversarial attack that an attacker can choose a similar domain and transfer the adversarial examples to the target domain and cause poor performance in target model. Based on domain adaptation theory, we then propose a defensive strategy, called Learn2Weight, which trains to predict the weight adjustments for a target model in order to defend against an attack of similar-domain adversarial examples. Using Amazon multi-domain sentiment classification datasets, we empirically show that Learn2Weight is effective against the attack compared to standard black-box defense methods such as adversarial training and defensive distillation. This work contributes to the growing literature on machine learning safety.
\end{abstract}

\section{Introduction}

As machine learning models are applied to more and more real-world tasks, addressing machine learning safety is becoming an increasingly pressing issue. Deep learning algorithms have been shown to be vulnerable to adversarial examples \citep{szegedy2013intriguing,goodfellow2014explaining,papernot2016transferability}. In particular, prior black-box adversarial attacks assume that the adversary is not aware of the target model architecture, parameters or training data, but is capable of querying the target model with supplied inputs and obtaining the output predictions. The phenomenon that adversarial examples generated from one model may also be adversarial to another model is known as adversarial transferability \citep{szegedy2013intriguing}. 

Motivated by adversarial transferability, we conjecture another black-box attack pipeline where the adversary does not even need to have access to the target model nor query labels from crafted inputs.  Instead, as long as the adversary knows the task of the target, they can choose a similar domain to build a substitute model, and then attack the target model with adversarial examples that are generated from the attack domain.

\begin{figure*}[htp]
  \centering
  \caption{Diagrammatic representation of the attack}
  \subfigure[Generalized architecture of similarity-based attacks.]{\hspace*{-0.2cm}\includegraphics[scale=0.2]{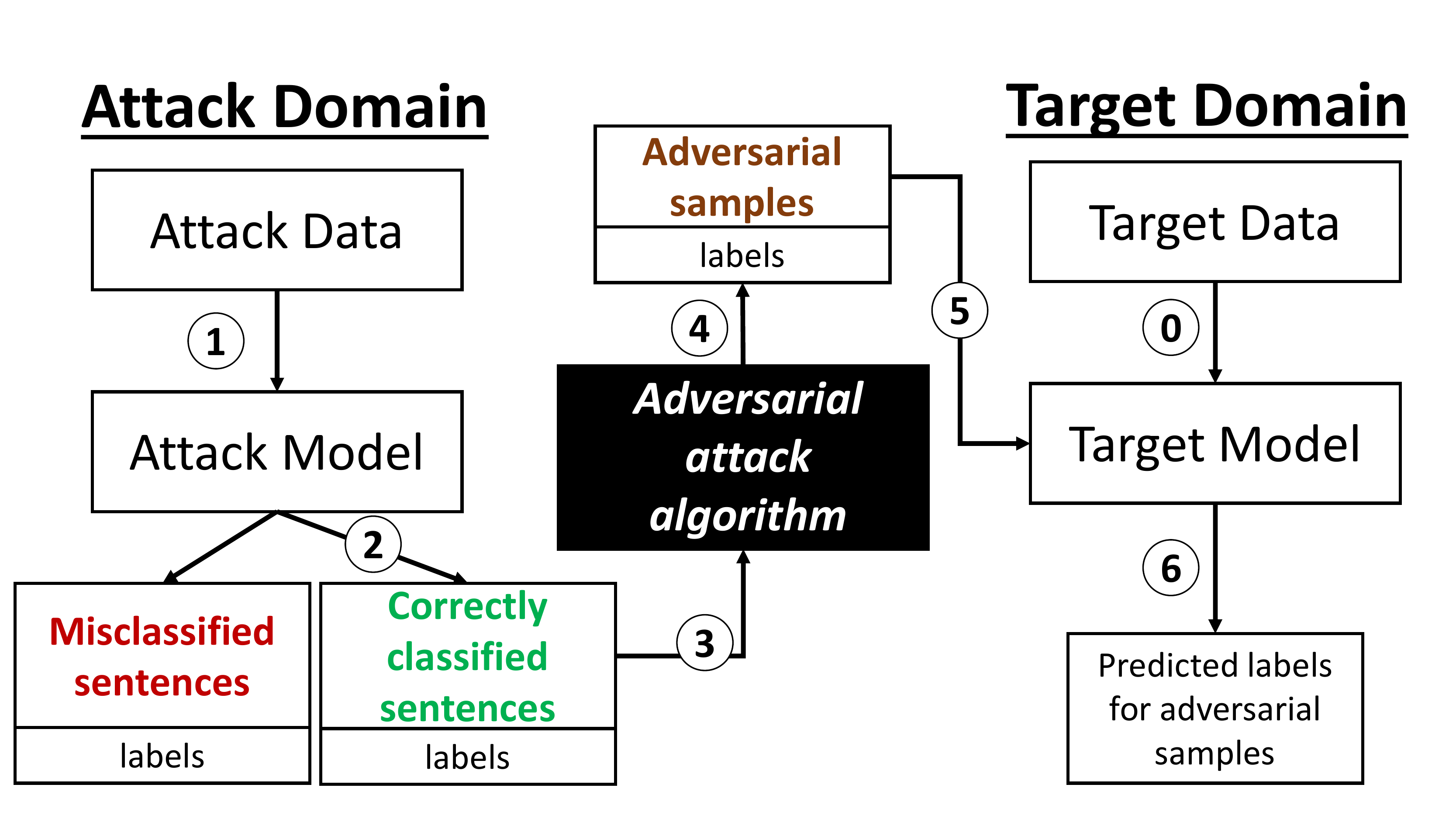}}\quad
  \subfigure[Flow of how an adversary physician can leverage similarity attack to fool opioid risk models.]{\hspace*{0.2cm}\includegraphics[scale=0.08]{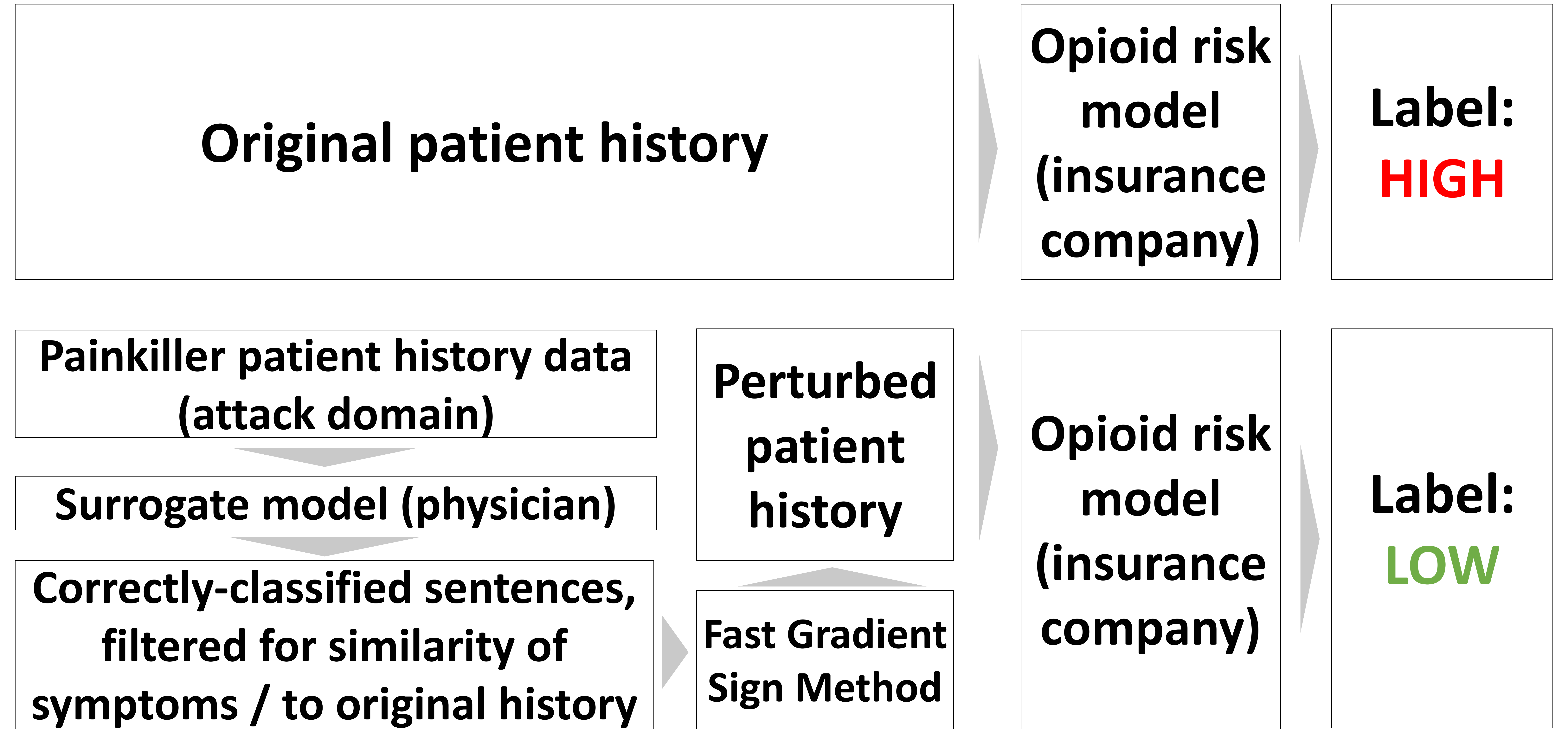}}
  \label{fig:insur}
\end{figure*}

The similar-domain adversarial attack may be more practical than prior blackbox attacks as label querying from the target model is not needed. This attack can be illustrated with the following example (Figure \ref{fig:insur}b) in medical insurance fraud \citep{Finlayson1287}.
Insurance companies may use hypothetical opioid risk models to classify the likelihood (high/low) of a patient to abuse the opioids to be prescribed, based on the patient's medical history as text input. 
Physicians can run the original patient history through the attack pipeline to generate an adversarial patient history, where the original is more likely to be rejected ("High" risk) and the adversarial is more likely to be accepted ("Low" risk). Perturbations in patient history could be, for example, a slight perturbation from "alcohol abuse" to "alcohol dependence", and it may successfully fool the insurance company's model. 

Based on domain adaption theory \citep{36364}, we conjecture that domain-variant features cause the success of the similar-domain attack. The adversarial examples with domain-variant features are likely to reside in the low-density regions (far away from decision boundary) of the empirical distribution of the target training data which could fool the target model \citep{zhang2019limitations}. 
Literature indicates that worsened generalizability is a tradeoff faced by existing defenses such as adversarial training \citep{raghunathan2019adversarial} and domain generalization techniques \citep{wang2019learning}. In trying to increase robustness against adversarial inputs, a model faces a tradeoff of weakened accuracy towards clean inputs. Given that an adversarial training loss function is composed of a loss against clean inputs and loss against adversarial inputs, improper optimization where the latter is highly-optimized and the former weakly-optimized does not improve general performance in the real-world. To curb this issue, methods have been proposed \citep{10.5555/3327345.3327409, zhang2019limitations, 10.1145/3338501.3357369}, such as factoring in under-represented data points in training set.

To defend against this similar-domain adversarial attack, we propose a meta learning approach, \textbf{Learn2Weight}, so that the target model's decision boundary can adapt to the examples from low-density regions.
Experiments confirm the effectiveness of our approach against the similar-domain attack over other baseline defense methods. Moreover, our approach is able to improve robustness accuracy without losing the target model's standard generalization accuracy.

\noindent
Our contribution can be summarized as follows
\footnote[2]{$\dagger$ indicates supplementary information can be found in the appendix (Appendix: \citet{https://doi.org/10.48550/arxiv.2205.07315}).}:
\begin{itemize}
\itemsep0em 
    \item We are among the first to demonstrate the similar-domain adversarial attack, leveraging domain adaptation to create adversarial perturbations that compromise NLP models. This attack pipeline relaxes the previous black-box attack assumption that the adversary has access to the target model and can query the model with crafted examples.
    \item We propose a defensive strategy for this attack based on domain adaptation theory and meta learning. Experiments show the effectiveness of our approach over existing defenses against the similar-domain adversarial attack.
\end{itemize}

\newpage

\section{Related Work}
\citet{zhang2020adversarial} provides a survey of adversarial attacks in NLP. Existing research proposes different attack methods for generating adversarial text examples \citep{moosavi2016deepfool,ebrahimi2018hotflip,wallace2019universal}.
The crafted adversarial text examples have been shown to fool state-of-the-art NLP systems, e.g. BERT \citep{jin2019bert}. A large body of adversarial attack research focuses on black-box attack where the adversary builds a substitute model by querying the target model with supplied inputs and obtaining the output predictions. The key idea behind such
black-box attack is that adversarial examples generated
from one model may also be misclassified
by another model, which is known as adversarial
transferability \citep{szegedy2013intriguing,NIPS2019_9275}. While prior work examines the transferability
between different models trained over the same
dataset, or the transferability between the same
or different models trained over disjoint subsets of
a dataset, our work examines the adversarial
transferability between different domains, which we call a similar-domain adversarial attack.

\begin{table*}[]
\begin{adjustwidth}{0cm}{}
\footnotesize 
\centering
\begin{tabular}{llc}
\hline
\multicolumn{1}{c}{} & \multicolumn{1}{c}{Attack domain: baby, Target domain: books} &  \\ \hline

\begin{tabular}[c]{@{}l@{}}Original sentence \\ (Actual label: {\color{cadmiumgreen}{Pos}})\end{tabular} & \begin{tabular}[c]{@{}l@{}}I purchased this toy for my son when he was 4 months old. At first, he seemed \\ a little intimidated by the toys.\end{tabular} & {\color{cadmiumgreen}{Pos (0.712)}} \\
Adversarial sentence & \textit{\begin{tabular}[c]{@{}l@{}}I {\color{blue}{obtained}} this toy{\color{blue}{s}} for my {\color{blue}{children}} when he was 4 {\color{blue}{weeks senior}}. At first, he \\ {\color{blue}{hoped}} a {\color{blue}{modest harassed}} by the {\color{blue}{toy}}.\end{tabular}} & {\color{red}{Neg (0.364)}} \\ \hline

\begin{tabular}[c]{@{}l@{}}Original sentence \\ (Actual label: {\color{cadmiumgreen}{Pos}})\end{tabular} & \begin{tabular}[c]{@{}l@{}}It felt like a big commitment for me to have to run the program 2 times a day, \\ and near the end of my pregnancy I was annoyed with having anything \\ strapped across my belly.\end{tabular} & {\color{cadmiumgreen}{Pos (0.825)}} \\
Adversarial sentence & \textit{\begin{tabular}[c]{@{}l@{}}It felt like a big {\color{blue}{committed}} for me to have to run the program 2 {\color{blue}{length}} a day, and \\ near the end of my pregnancy I was annoyed with {\color{blue}{takes}} anything strapped \\ across my belly.\end{tabular}} & {\color{red}{Neg (0.420)}} \\ \hline

\multicolumn{1}{c}{} & \multicolumn{1}{c}{Attack domain: dvd, Target domain: baby} &  \\ \hline

\begin{tabular}[c]{@{}l@{}}Original sentence \\ (Actual label: {\color{cadmiumgreen}{Pos}})\end{tabular} & \begin{tabular}[c]{@{}l@{}}Fast times at ridgemont high is a clever, insightful, and wicked film! It is not \\ just another teen movie.\end{tabular} & {\color{cadmiumgreen}{Pos (0.614)}} \\
Adversarial sentence & \textit{\begin{tabular}[c]{@{}l@{}}{\color{blue}{Sooner days}} at ridgemont high is a {\color{blue}{sane}}, {\color{blue}{thoughtful}}, and wicked {\color{blue}{flick}}! It is not \\ just another {\color{blue}{adolescent flick}}.\end{tabular}} & {\color{red}{Neg (0.335)}} \\ \hline

\begin{tabular}[c]{@{}l@{}}Original sentence \\ (Actual label: {\color{cadmiumgreen}{Pos}})\end{tabular} & \begin{tabular}[c]{@{}l@{}}This dvd gives a very good 60 minute workout. As others have pointed out the \\ cardio is very dancy. The first time I did it, I felt a bit awkward with the steps.\end{tabular} & {\color{cadmiumgreen}{Pos (0.647)}} \\
Adversarial sentence & \textit{\begin{tabular}[c]{@{}l@{}}This dvd gives a {\color{blue}{awfully okay}} 60 minute {\color{blue}{exercise}}. As others have pointed out the \\ cardio is very dancy. The first time I did it, I {\color{blue}{perceived}} a bit awkward with the steps.\end{tabular}} & {\color{red}{Neg (0.258)}} \\ \hline

\end{tabular}
\caption{Comparison of attack domain sentences correctly classified when unperturbed by respective attack domain models and target domain models, then misclassified after perturbation by target models trained on \textbf{books} and \textbf{baby} domain. The {\color{blue}{perturbations}} are in blue, and prediction confidence in brackets.}
\label{tab:emp.examples}
\end{adjustwidth}
\end{table*}

\section{Similar-domain Adversarial Attack}
\label{sec:cd}

\subsection{Adversarial attack background}
Adversarial attacks modify inputs to cause errors in machine learning inference \citep{szegedy2013intriguing}.
We use the basic gradient-based attack method \textit{Fast Gradient Sign Method (FGSM)} \citep{goodfellow2014explaining}, with perturbation rate $\varepsilon=0.4$. 
Other NLP adversarial generation algorithms could also be used, such as
\textit{Rand-FGSM} \citep{tramer2017ensemble}, 
\textit{Basic Iterative Method} \citep{kurakin2016adversarial, kurakin2016adversarial2, xie2018improving},
DeepFool \citep{moosavi2016deepfool}, HotFlip \citep{ebrahimi2018hotflip}, universal adversarial trigger \citep{wallace2019universal}, and TextFooler \citep{jin2019bert}.
To perform gradient-based perturbations upon discrete space data, we follow \citet{10.1109/MILCOM.2016.7795300} to generate adversarial text.
Our proposed similar-domain adversarial attack is in-variant to adversarial algorithm, meaning that the adversarial algorithm used would not affect the attack performance. 

\begin{definition}{\textbf{NLP Adversarial Generation.}}
We denote $\texttt{Adv}(\theta; \mathbf{x}; \varepsilon)$ as an NLP adversarial generation method.
The goal of $\texttt{Adv}$ is to maximize the misclassification rate on perturbed inputs:
\newline
$\mathbf{x}^{\textnormal{adv}} = \texttt{Adv}(\theta; \mathbf{x})$ \textit{s.t.} $y \neq \mathscr{f}(\theta; \mathbf{x}^{\textnormal{adv}})$.
\end{definition}

\subsection{Similar-domain Adversarial attack } 
We present the architecture of similar-domain adversarial attack in Figure \ref{fig:insur}a. 
The defender, the target of the attack, constructs a target model (parameters $\theta_i$) trained on domain text data $X_i$ \circled{0}. 
An attacker, only having a rough idea about the target's task but lacking direct access to the target data or target model parameters, collects attack data from a similar domain $X_j \sim \mathcal{X}$ and trains an attack model (parameters $\theta_j$) \circled{1}. 
They run the attack model on the test data \circled{2} to obtain correctly-classified instances \circled{3}. 
They chooses an adversarial attack algorithm and generate a set of adversarial samples $X_j^{\textnormal{adv}}$ \circled{4}. 
They expose $X_j^{\textnormal{adv}}$ to the target model, hoping $X_j^{\textnormal{adv}}$ misleads the target model to produce an output of their choice \circled{5}. 
The attacker's objective is to maximize the misclassification per label and minimize the accuracy w.r.t. perturbed inputs ($\max$ Eqt \ref{equation:js}), 
while the defender's objective is to maximize the accuracy w.r.t. perturbed inputs ($\min$ Eqt \ref{equation:js}).
This type of attack works best as an adversarial attack that compromises systems that base decision-making on one-instance.

\vspace{-0.4cm}
\begin{equation}
\small
\begin{split}
\mathbb{E}_{\mathbf{x}_j, y_j \sim X_j, Y_j}[\mathscr{f}(\theta_i; \texttt{Adv}(\theta_j; \mathbf{x}_j))-y_j]
\end{split}
\label{equation:js}
\end{equation}

\begin{definition}{\textbf{Similar-domain Adversarial Attack.}}
Target model $\mathscr{f}$, trained on target domain data $X_i$, is a deep neural network model with weights $\theta_i$ mapping text instances to labels: 
$Y_i = \mathscr{f}(\theta_i; X_i)$.
An adversary chooses source attack domain $X_j$, builds substitute model 
$\mathscr{f}(\theta_j; X_j)$,
and generates a set of adversarial examples $X_j^{\textnormal{adv}}$ from $X_j$ using 
$\texttt{Adv}(\theta_j; X_j)$,
such that  during an attack 
\newline
$\mathscr{f}(\theta_i; X_j^{\textnormal{adv}}) = \mathscr{f}(\theta_j; X_j^{\textnormal{adv}})$.
\end{definition}

\begin{table*}[t]
\centering
\parbox{\textwidth}{
\centering
\resizebox{\textwidth}{!}{
\begin{tabular}{|l|ccc|cccc|cccc|ccc}
\hline
Target Domain & \multicolumn{4}{c}{book} & \multicolumn{3}{c}{magazine} & \multicolumn{4}{c|}{baby} \\ \hline
Original Accuracy 
& \multicolumn{4}{c}{0.880} & \multicolumn{3}{c}{0.960} & \multicolumn{4}{c|}{0.890} \\
Intra-attack Accuracy 
& \multicolumn{4}{c}{0.525} & \multicolumn{3}{c}{0.570} & \multicolumn{4}{c|}{0.632} \\ \hline \hline
Attack Domain & magazine & baby & dvd & & baby & dvd & book & &  dvd & book & magazine \\ \hline
Unperturbed Accuracy 
& 0.745 
& 0.726 
& 0.646 
& 
& 0.673 
& 0.663 
& 0.739 
& 
& 0.652 
& 0.624 
& 0.665 
\\
After-attack Accuracy 
& 0.395 
& 0.398 
& 0.421 
& 
& {0.343} 
& {0.366} 
& 0.381 
& 
& {0.386} 
& {0.365} 
& {0.401} 
\\
SharedVocab 
& 0.455 
& 0.381 
& 0.255 
& 
& 0.381  
& 0.345 
& 0.260 
& 
& 0.255 
& 0.270 
& 0.260 
\\
Transfer Loss 
& 0.000 
& 0.017 
& {0.071} 
& 
& 0.010 
& 0.022 
& {0.079} 
& 
& 0.050 
& 0.066 
& {0.069}  
\\ \hline
\end{tabular}
}
}
\caption{
\textit{Domain shift \& similarity:}
Sorted in descending order of domain similarity,
we observe a lower after-attack accuracy when domain similarity increases.
}
\label{tab:empirical.attack}
\end{table*}

\section{Is the Attack Effective?}
\subsection{Setup}

\noindent\textbf{(Datasets)}
We sample domains from 
25 domain datasets, 
each containing 1,000 positive and 1,000 negative reviews for an Amazon product category,
sourced from the Amazon multi-domain sentiment classification benchmark
\citep{blitzer2007biographies}.

\noindent\textbf{(Models)} 
We evaluated our setup on several architectures commonly-used for sentiment classification, including 
LSTM \citep{wang-etal-2018-lstm}, 
GRU, 
BERT \citep{devlin-etal-2019-bert}, 
CNN \citep{kim2014convolutional}, 
and Logistic Regression \citep{Maas:2011:LWV:2002472.2002491}.

\noindent\textbf{(Domain similarity)}
refers to the similarity between attacker's chosen domain and defender's domain. \textbf{SharedVocab} measures the overlap of unique words, in each of the datasets; a higher degree of overlapping vocabulary implies the two domains are more similar. We also use \textbf{Transfer Loss}, a standard metric for domain adaptation \cite{blitzer2007biographies,Glorot:2011:DAL:3104482.3104547}, to measure domain similarity; lower loss indicates higher similarity. The test error from a target model trained on target domain $X_i$ and evaluated on attack domain $X_j$ returns transfer error {\small $e(X_j, X_i)$}. The baseline error {\small $e(X_i, X_i)$} term is the test error obtained from target model trained on target domain (train) data $X_i$ and tested on target domain (evaluation) data $X_i$. This computes the transfer loss, {\small $tf(X_j,X_i)=e(X_j,X_i)-e(X_i,X_i)$}.

\noindent\textbf{(Accuracy)} We first report the  accuracy of the target models on the target domain test samples before the attack as the \textit{original accuracy}. Then we measure the accuracy of the target models against adversarial samples crafted from the attack domain samples, denoted as the \textit{after-attack accuracy}. \textit{Intra-attack accuracy} denotes the after-attack accuracy where the attack domain is identical to the target domain. By comparing original and after-attack accuracy, we can evaluate the success of the attack. The greater the gap between the original and after-attack accuracy, the more successful the attack. \textit{Unperturbed accuracy} measures the accuracy of the target model against the complete, unperturbed test set of the attack domain, to demonstrate that any drop in classification accuracy is not from domain shift alone but from adversarial transferability.

\newpage
\subsection{Results}
The similar-domain adversarial attack results are presented in Table \ref{tab:empirical.attack}.  We see a significant gap between original accuracy and after-attack accuracy, indicating that this attack can impose a valid threat to a target NLP system. After the similar-domain adversarial attack, the accuracy drops dramatically by a large margin. Take the book target domain as an example: when the attack domain is magazine, the after-attack accuracy drops to 0.398, and when the attack domain is baby, the accuracy is 0.421. Moreover, we observe a positive correlation between transfer loss and after-attack accuracy, and a negative correlation between shared vocab and after-attack accuracy. 

\section{Defending Against Similar-domain Adversarial Attack}

In order to defend against a similarity based adversarial attack, it is critical to block adversarial transferability. Adversarial training is the most intuitive yet effective defense strategy for adversarial attack \citep{goodfellow2014explaining,madry2017towards}. However, this may not be effective for two reasons. First, there is no formal guidance for generating similar-domain adversarial examples because the defender has no idea what the attack data domain is. Second, simply feeding the target model with adversarial examples may even hurt the generalization of the target model \citep{su2018robustness,raghunathan2019adversarial,zhang2019theoretically}, which is also confirmed in our experiments. 

\subsection{Parameter Adaptation}
Meta learning techniques that modify parameters
\citep{ha2016hypernetworks,hu2018learning,kuen2019scaling} are concerned with adapting weights from one model into another, and generating/predicting the complete set of weights for a model given the input samples. In our context, distinctly different weights are produced for target models trained on inputs of different domains, and feature transferability \citep{10.5555/2969033.2969197} in the input space can be expected to translate to weights transferability in the parameter space. Rather than completely regenerating classification weights, our model robustification defense, \textit{Learn2Weight}, predicts the perturbation to existing weights 
$\theta^{*} = \theta_i + \widehat{\Delta \theta}$
for each new instance.

\subsection{Learn2Weight (L2W)\textsuperscript{$\dagger$}}
\label{sec:l2w}

We conjecture that an effective defense strategy is to perturb the target model weights depending on the feature distribution of the input instance. 
In inference (Algorithm \ref{alg:inf}),
L2W recalculates the target model weights depending on the input. 
During training (Algorithm \ref{alg:train}),
L2W trains on sentences from different  domains and a weight differential for that domain (the weight adjustment required to tune the target model's weights to adapt to the input's domain). We obtain the weight differential $\Delta \theta$ by finding the difference between the weights $\theta_j$ trained on sentence:label pairs from a specific domain 
$X_j \sim \mathcal{X}$
and weights $\theta_i$ trained on sentence:label pairs from the target domain $X_i$. Other training models may be possible; here we trained a sequence-to-sequence network \citep{10.5555/2969033.2969173} on sentence:$\Delta \theta$ pairs.

\subsection{Perturbation Sets Generation\textsuperscript{$\dagger$}}

To generate synthetic domains of varying domain similarity $\mathbf{S} = \{X_j:Y_j\}_{j=1}^T$ so that defenders defend their model using only target domain data $X_i$, 
a defender iteratively generates perturbation sets that 
minimizes transfer loss while maximizing adversarial perturbations (Algorithm \ref{alg:tfo}).
A \textit{perturbation set} is a set containing subsets of perturbed inputs \citep{alzantot-etal-2018-generating, DBLP:conf/icml/WongSK19}.
To construct one perturbation set (Eqt \ref{equation:perturbation_set}), we utilize an iterative minimax algorithm, 
where we iteratively apply a maximizing adversarial perturbation factor $\varepsilon \geq \varepsilon_{\textnormal{min}}$, 
and accept the batch of perturbed inputs if it yields a minimizing input distance $\texttt{dist} \leq d_{\textnormal{max}}$.
We repeat this $T$ times. 
We use transfer loss as the distance metric
to optimize for domain similarity.
We retain FGSM as the adversarial attack algorithm.

\vspace{-0.3cm}
\begin{equation}
\small
\begin{split}
X^{*} & := \min \texttt{dist}(X^{*}, X_i) \leq d_{\textnormal{max}} \\
X^{*} & := \min \argmax_{\varepsilon \sim [\varepsilon_{\textnormal{min}}, 1]} \texttt{dist}(\texttt{Adv}(\theta_i; X_i; \varepsilon), X_i) \\
X^{*} & := \min \argmax_{\varepsilon \sim [\varepsilon_{\textnormal{min}}, 1]} [e(\texttt{Adv}(\theta_i; X_i; \varepsilon), X_i) - e(X_i, X_i)] 
\end{split}
\label{equation:perturbation_set}
\end{equation}

{
\newpage
\begin{algorithm}[H]
  \footnotesize 
  \caption{Learn2Weight (Inference)}
  \SetKwInOut{Input}{Input}
  \SetKwInOut{Output}{Output}
  \SetKwProg{inference}{inference}{}{}

  \inference{$(X_j^{\textnormal{adv}}, \mathscr{h}(\theta^{\mathscr{mf}}), \mathscr{f}(\theta_i)$)}{
    \Input{test-time inputs $X_j^{\textnormal{adv}}$; L2W $\mathscr{h}(\theta^{\mathscr{mf}})$; \\base learner $\mathscr{f}(\theta_i)$}
    \Output{label $\hat{y}$}
    \medbreak
    Compute parameter differential w.r.t. $X_j^{\textnormal{adv}}$.\\
    $\widehat{\Delta \theta} \gets \mathscr{h}(\theta^{\mathscr{mf}}; X_j^{\textnormal{adv}})$\;
    \medbreak
    Update $\theta^{\mathscr{f}}$.\\
    $\hat{y} \gets \mathscr{f}(\theta_i + \widehat{\Delta \theta}; X_j^{\textnormal{adv}})$\;
    \medbreak
    \KwRet{$\hat{y}$}\;
  }
  \label{alg:inf}
\end{algorithm}
\vspace{-0.8cm}
\begin{algorithm}[H]
  \footnotesize 
  \caption{Learn2Weight (Training)}
  \SetKwInOut{Input}{Input}
  \SetKwInOut{Output}{Output}
  \SetKwProg{train}{train}{}{}
  \train{$(\mathbf{S}, \mathbf{D}, \theta_i, \mathbf{E}^{\mathscr{f}}, \mathbf{E}^{\mathscr{mf}}$)}{
    \Input{ domains (perturbation sets) $\mathbf{S}$, target domain $\mathbf{D}=\{X_i:Y_i\}$, base learner parameters $\theta_i$, epochs $\mathbf{E}^{\mathscr{f}}$ \& $\mathbf{E}^{\mathscr{mf}}$}
    \Output{L2W parameters $\theta^{\mathscr{mf}}$}
    \medbreak
    Initialize empty set $\Theta$ to store parameter differential. \\
    $\Theta \gets \emptyset$;\;
    \medbreak
    Compute $X_j \mapsto \Delta \theta$. \\
    \ForEach{$X_j:Y_j \in (\mathbf{D} \cup \mathbf{S})$}{%
        \For{$e \gets 0$ \KwTo $\mathbf{E}^{\mathscr{f}}$}{%
            $\theta_{j, e}^{\mathscr{f}} := \theta_{j, e-1}^{\mathscr{f}} - \sum_{\mathbf{x}, y}^{X_j, Y_j} \frac{\partial \mathcal{L}(\mathbf{x}, y)}{\partial \theta^{\mathscr{f}}}$\;\\

        }
        $\Delta \theta \gets \theta_j^{\mathscr{f}}-\theta_i$\;\\
        $\Theta \gets \Delta \theta$;\; 
    }
    \medbreak
    Compute $\theta^{\mathscr{mf}}$.\\
    \For{$e \gets 0$ \KwTo $\mathbf{E}^{\mathscr{mf}}$}{%
        $\theta_{e}^{\mathscr{mf}} := \theta_{e-1}^{\mathscr{mf}} - \sum_{X_j, \Delta \theta}^{(X_i \cup \mathbf{S}), \Theta} \frac{\partial \mathcal{L}(X_j, \Delta \theta)}{\partial \theta^{\mathscr{mf}}}$\;\\
    }
    \KwRet{$\theta^{\mathscr{mf}}$}\; 
  }
  \label{alg:train}
\end{algorithm}
\vspace{-0.8cm}
\begin{algorithm}[H]
  \footnotesize 
  \caption{Perturbation Sets Generation}
  \SetKwInOut{Input}{Input}
  \SetKwInOut{Output}{Output}
  \SetKwProg{PerturbationSet}{PerturbationSet}{}{}
  
  \PerturbationSet{$(\mathbf{D}, \theta_i; T, R; \texttt{dist}, d_{\textnormal{max}}; \varepsilon, \gamma)$}{
    \Input{target domain $\mathbf{D}=\{X_i:Y_i\}$, parameters $\theta_i$; number of perturbation sets $T=10$, max iterations $R = 10$; distance metric $\texttt{dist} = tf(X_i, X_j)$, max distance $d_{\textnormal{max}} = 0.1$; initial perturbation rate $\varepsilon = 0.9$, perturbation learning rate $\gamma = 0.05$; }
    \Output{set $\mathbf{S}$ containing $T$ perturbation sets}
    \medbreak
    Initialize empty $\mathbf{S}$ to store perturbation sets $S_t$. \\
    $\mathbf{S} \gets \emptyset$;
    \medbreak
    \While {$t < T$}{
    
        Run next iteration $r$ until $S_t$ meets conditions. \\
        \For {$r\gets0$ \KwTo $R$}{
            Apply adversarial perturbations to $X$. \\
            $S_{t, r} \gets \texttt{Adv}(\theta_i; X_i; \varepsilon)$;
            \medbreak
            Evaluate distance conditions. \\
            \If {$\texttt{dist}(S_{t, r}, X_i) \leq d_{\textnormal{max}}$}{
                \If {$\sigma^2(\mathbf{S} \cup S_{t, r}) > \sigma^2(\mathbf{S})$}{
                    $\mathbf{S} \gets \{S_{t, r}:Y_i\}$;\\
                    \textbf{continue}; \\
                }
                }
            \Else {
                Adjust hyperparameters. \\
                $\varepsilon \gets \varepsilon - \gamma$; }
            
        }
        
        $t \gets t+1$;
    }
    
    \medbreak
    \KwRet{$\mathbf{S}$}\; 
  }
\label{alg:tfo}
\end{algorithm}
}

\subsection{Explanation: Blocking Transferability}

To facilitate our explanation, we adapt from domain adaptation literature \citep{36364, pmlr-v97-liu19b, pmlr-v97-zhang19i}:

\vspace{-0.4cm}
\begin{equation}
\small
\begin{split}
e(X_j^{\textnormal{adv}}, X_i) \leq e(X_i, X_i) + d_{\mathcal{H} \Delta \mathcal{H}} (X_j^{\textnormal{adv}}, X_i) + \lambda
\end{split}
\label{equation:dd0}
\end{equation}

where $\mathcal{H}$ is the hypothesis space, $h$ is a hypothesis function that returns labels $\{ 0,1 \}$, and $e(X_i, X_i)$ and $e(X_j^{\textnormal{adv}}, X_i)$ are the generalization errors from passing target domain data $X_i$ and adversarial data $X_j^{\textnormal{adv}}$ through a classifier trained on $X_i$. $d_{\mathcal{H} \Delta \mathcal{H}} (X_j^{\textnormal{adv}}, X_i)$ is the $\mathcal{H} \Delta \mathcal{H}$-distance between $X_i$ and $X_j^{\textnormal{adv}}$, and measures the divergence between the feature distributions of $X_j^{\textnormal{adv}}$ and $X_i$. $e_{X_j^{\textnormal{adv}}} (h, h^{'})$ and $e_{X_i} (h, h^{'})$ represent the probability that $h$ disagrees with $h^{'}$ on the label of an input in the domain space $X_j^{\textnormal{adv}}$ and $X_i$ respectively.

\vspace{-0.4cm}
\begin{equation}
\small
\begin{split}
d_{\mathcal{H} \Delta \mathcal{H}} (X_j^{\textnormal{adv}}, X_i) & = \mathop{\sup}_{h, h^{'} \in \mathcal{H}} | e_{X_j^{\textnormal{adv}}} (h, h^{'}) - e_{X_i} (h, h^{'}) | \\
d_{\mathcal{H} \Delta \mathcal{H}} (X_j^{\textnormal{adv}}, X_i) & = \mathop{\sup}_{h, h^{'} \in \mathcal{H}} \Big| \mathbb{E}_{x_j \sim X_j} \big[ | (h(x_j) - h^{'}(x_j) | \big] \Big| \\ 
& \phantom{=} - \Big| \mathbb{E}_{x_i \sim X_i} \big[ | (h(x_i) - h^{'}(x_i) | \big] \Big|
\end{split}
\label{equation:dd0}
\end{equation}

Divergence $d_{\mathcal{H} \Delta \mathcal{H}}$ measures the divergence between feature distributions $X_j^{\textnormal{adv}}$ and $X_i$. Higher $d_{\mathcal{H} \Delta \mathcal{H}}$ indicates less shared features between 2 domains. The greater the intersection between feature distributions, the greater the proportion of domain-variant features; one approach to domain adaptation is learning domain-invariant features representations \citep{pmlr-v97-zhao19a} to minimize $d_{\mathcal{H} \Delta \mathcal{H}}$.

\noindent\textbf{Explaining similarity-domain attacks.} As demonstrated by empirical results, $e(X_j^{\textnormal{adv}}, X_i)$ increases in a similarity-based attack setting, and this would arise if $d_{\mathcal{H} \Delta \mathcal{H}}$ increases correspondingly. $d_{\mathcal{H} \Delta \mathcal{H}}$ computes inconsistent labels from inconsistent feature distributions, and attributes the success of the attack to domain-variant features. 

\textit{FGSM} and variants adjust the input data to maximize the loss based on the backpropagated gradients of a model trained on $X_j$. As our pipeline used correctly-labelled sentences before adversarially perturbing them, we can infer that perturbations applied to $X_j$ were not class-dependent (i.e. the success of the attack is not based on the removal of class-specific features), but class-independent features. It is already difficult for a model trained on $X_j$ to classify when there is insufficient class-dependent features (hence a high $tf(X_j^{\textnormal{adv}},X_i)$); in a cross-domain setting, it must be even more difficult for a model trained on $X_i$ to classify given a shortage of domain-invariant, class-dependent features.

\vspace{-0.5cm}
\begin{equation}
\small
\begin{split}
d_{\mathcal{H} \Delta \mathcal{H}} & \geq e(X_j^{\textnormal{adv}}, X_i) - e(X_i, X_i) - \lambda \\
d_{\mathcal{H} \Delta \mathcal{H}} & \geq tf(X_j^{\textnormal{adv}}, X_i) - \lambda
\end{split}
\label{equation:dd0}
\end{equation}

\noindent\textbf{Explaining Learn2Weight.} L2W minimizes divergence by training on {\small 
$\{ d_{\mathcal{H} \Delta \mathcal{H}} (X_j, X_i) : \Delta \theta \}$ } pairs,
such that {\small $\Delta \theta = L2W(d_{\mathcal{H} \Delta \mathcal{H}}(X_j, X_i))$ },
where {\small $d_{\mathcal{H} \Delta \mathcal{H}} (X_j, X_i)$ } is reconstructed from the difference between $X_j$ and $X_i$. 
The target model possesses a decision boundary \citep{pmlr-v97-liu19b} to classify inputs based on whether they cross the boundary or not; adversarial inputs have a tendency of being near the boundary and fooling it. 
Meta learning applies perturbations to the decision boundary such that the boundary covers certain adversarial inputs otherwise misclassified, and in this way blocks transferability. The advantage of training on multiple domains $\{X_{j}\}^{T}_{j=1}$ is that the after-L2W divergence between $X_j^{\textnormal{adv}}$ and $X_i$ is smaller because L2W's weight perturbations render the decision boundary more precise in classifying inputs. 

\noindent\textbf{Explaining perturbation sets.} We attributed why adversarial sentences $X_j^{\textnormal{adv}}$ are computed to be domain-dissimilar despite originating from $X_j$ due to insufficient domain-invariant, class-dependent features resulting in low $e(X_j^{\textnormal{adv}},X_i)$, i.e. low $tf(X_j^{\textnormal{adv}},X_i)$. To replicate this phenomenon in natural domains, we iteratively perturb $X_i$ to increase the proportion of class-independent features. This approximates the real-world similarity-based attack scenario where class-dependent features may be limited for inference. By generating the synthetic data, we are feeding L2W attack data with variations in $d_{\mathcal{H} \Delta \mathcal{H}}$ and class-independent feature distributions. This prepares L2W to robustify weights $\theta_i$ when such feature distributions are met.

\begin{table*}[t]
\centering
\parbox{\textwidth}{
\centering
\resizebox{\textwidth}{!}{
\begin{tabular}{|l|cccccccccccc|}
\hline
Target Domain & \multicolumn{8}{c}{{{magazine}}} & \multicolumn{4}{c|}{{{baby}}} \\ \hline
Attack Domain &&& & {{baby}} & {{dvd}} & {{book}} &&& & {{dvd}} & {{book}} & {{magazine}} \\ \hline
After-attack Accuracy
&&& & 0.381 & 0.366 & 0.343 &&& & 0.365 & 0.386 & 0.401 \\ \hline \hline
After-defense Accuracy
&&& & &  & &&& & & & \\ \hline
Adversarial Training &&& & 0.639 & 0.559 & 0.657 &&& & 0.558 & 0.577 & 0.661 \\ 
Defensive Distillation &&& & 0.549 & 0.561 & 0.597 & &&& 0.588 & 0.629 & 0.577 \\ 
Perturbation Sets Adversarial Training &&& & 0.608 & 0.637 & 0.620 &&& & 0.604 & 0.620 & 0.587 \\
\textbf{Learn2Weight} &&& & \textbf{0.796} & \textbf{0.842} &\textbf{ 0.843}& &&& \textbf{0.774}&\textbf{ 0.751}& \textbf{0.737} \\ \hline
\end{tabular}
}
}
\caption{
\textit{After-defense Accuracy: }
Learn2Weight outperforms the baseline and ablation methods.
}
\label{tab:defenses}
\end{table*}
\begin{table*}[t]
\centering
\parbox{\textwidth}{
\centering
\resizebox{\textwidth}{!}{
\begin{tabular}{|clcccccccccc|}
\hline
\multicolumn{1}{|c}{\begin{tabular}[c]{@{}c@{}}Target \\ Domain\end{tabular}} &
\multicolumn{1}{c}{\begin{tabular}[c]{@{}c@{}}Attack \\ Domain\end{tabular}} & \multicolumn{5}{c}{
After-Attack Accuracy
} & \multicolumn{5}{c|}{ After-Defense Accuracy} \\ \hline
\multicolumn{2}{|c|}{} & BERT & LSTM & GRU & CNN & \multicolumn{1}{c|}{LogReg} & {BERT} & {LSTM} & {GRU} & {CNN} & {LogReg} \\ \hline
\multicolumn{1}{|c}{\multirow{3}{*}{{{book}}}} 
& \multicolumn{1}{|l|}{{{dvd}}} & 0.342 & 0.413 & 0.477 & 0.335 & \multicolumn{1}{c|}{0.440} & {0.786} & {0.847} & {0.804} & {0.816} & {0.782} \\
& \multicolumn{1}{|l|}{{{kitchenware}}} & 0.350 & 0.372 & 0.325 & 0.353 & \multicolumn{1}{c|}{0.425} & {0.765} & {0.826} & {0.795} & {0.742} & {0.767} \\
& \multicolumn{1}{|l|}{{{electronics}}} & 0.400  & 0.389 & 0.416 & 0.315 & \multicolumn{1}{c|}{0.460} & {0.792} & {0.812} & {0.784} & {0.770} & {0.725} \\ \hline
\multicolumn{1}{|c}{\multirow{3}{*}{{{dvd}}}} 
& \multicolumn{1}{|l|}{{{book}}} & 0.326  & 0.434 & 0.479 & 0.383 & \multicolumn{1}{c|}{0.490} & {0.816} & {0.795} & {0.824} & {0.804} & {0.794} \\
& \multicolumn{1}{|l|}{{{kitchenware}}} & 0.355  & 0.370 & 0.379 & 0.359 & \multicolumn{1}{c|}{0.490} & {0.728} & {0.796} & {0.755} & {0.735} & {0.695} \\
& \multicolumn{1}{|l|}{{{electronics}}} & 0.387  & 0.377 & 0.332 & 0.348 & \multicolumn{1}{c|}{0.455} & {0.825} & {0.836} & {0.812} & {0.834} & {0.796} \\ \hline
\multicolumn{1}{|c}{\multirow{3}{*}{{{electronics}}}} 
& \multicolumn{1}{|l|}{{{book}}} & 0.425  & 0.394 & 0.473 & 0.358 & \multicolumn{1}{c|}{0.474} & {0.775} & {0.821} & {0.795} & {0.782} & {0.712} \\
& \multicolumn{1}{|l|}{{{dvd}}} & 0.342  & 0.395 & 0.452 & 0.368 & \multicolumn{1}{c|}{0.493} & {0.784} & {0.845} & {0.855} & {0.842} & {0.792} \\
& \multicolumn{1}{|l|}{{{kitchenware}}} & 0.390  & 0.384 & 0.464 & 0.329 & \multicolumn{1}{c|}{0.432} & {0.730} & {0.824} & {0.753} & {0.724} & {0.678} \\ \hline
\end{tabular}
}
}
\caption{
\textit{Models: }
L2W retains high after-defense accuracy at varying attack model architectures.
}
\label{tab:defenses2}
\end{table*}

\newpage
\section{Experiments\textsuperscript{$\dagger$}}

\subsection{Baselines}

\noindent\textbf{Defensive distillation \citep{papernot2016distillation, 10.1145/3052973.3053009}:} The high-level implementation of \textit{defensive distillation} is to first train an initial model against target domain inputs and labels, and retrieve the raw class probability scores. 
The predicted probability values would be used as the new labels for the same target sentences, and we would train a new model based on this new label-sentence pair.

\noindent\textbf{Adversarial training \citep{goodfellow2014explaining,madry2017towards}:} It is shown that injecting adversarial examples throughout training increases the robustness of target neural network models. In this baseline, target model is trained with both original training data and adversarial examples generated from original training data. However, since the adversarial examples are still generated from the target domain, it is unlikely that the method can defend against a similar-domain adversarial attack, which is the result of domain-variant features.   

\noindent\textbf{Perturbation sets adversarial training: }
This ablation baseline tests for incremental performance to a baseline defense using domain-variant inputs. 
We adapt adversarial training to be trained on 
perturbation sets (synthetic domains) generated with Algorithm \ref{alg:tfo}
with respect to target domain $X_i$.

\subsection{Learn2Weight Performance}
\label{sec:res}
 
\noindent \textbf{Defense performance}. We present the results of different defense baselines in Table \ref{tab:defenses}. First, we can see that L2W achieves the highest after-defense accuracy against the adversarial attack. Take the \textit{magazine} as target domain for example: if the adversary chooses to use \textit{book} data as the attack domain, it would reduce the target model accuracy to 0.343. However, L2W can improve the performance to 0.843, which is a significant and substantial improvement against the attack. This improvement also exist across different target/attack domain pairs. Second, we see that all defense methods can improve the accuracy to some extent which indicates the importance and effectiveness of having robust training for machine learning models. 

\noindent\textbf{Attack model architectures.} So far, all the results are conducted using the same LSTM as the target/attack model.
Here, we keep the target model unchanged, but vary the architecture of the attack model for the generation of adversarial examples. 
LSTM (GRU) is configured with
64 cells, 
tokens embedded with respect to GloVe,
$\texttt{sigmoid}$ ($\texttt{tanh}$) activation function,
randomly-initialized and trained with Adam optimizer and $80\%$ ($60\%$) dropout, based on \citet{wang-etal-2018-lstm}.
CNN is configured with 
accepting tokens embedded with respect to GloVe \citep{pennington2014glove}, 
3 convolutional layers with kernel widths of 3, 4, and 5, all with 100 output channels,
and randomly-initialized, based on \citet{kim2014convolutional}.
We configure Logistic Regression based on \citet{Maas:2011:LWV:2002472.2002491}.
Based on \citet{devlin-etal-2019-bert}, we initialize a pretrained BERT with its own embeddings.
Models are trained until reaching state-of-the-art validation accuracy (early-stopping pauses training at loss 0.5).

We present the results of different attack model architectures in Table \ref{tab:defenses2}. First, the similar-domain adversarial attack is model-agnostic and it does not require the target and attack model to have identical architectures. We can see that all four attack model architectures are able to reduce the target model accuracy. Second, the results suggest that L2W is also model-agnostic as it can substantially improve the after-defense accuracy regardless which attack model is used. 

\section{Conclusion}
In this newly-proposed, empirically-effective similar-domain adversarial attack, an adversary can choose a similar domain to the target task, build a substitute model and produce adversarial examples to fool the target model. We also propose a defense strategy, Learn2Weight, that learns to adapt the target model's weight using crafted adversarial examples. Compared with other adversarial defense strategies, Learn2Weight can improve the target model robustness against the similar-domain attack. 
Our method demonstrates properties of a good adversarial defense, such as adopting a defense architecture that adapts to situations/inputs rather than compromising standard error versus robustness error, to leverage class-independent properties in domain-variant text, and factoring in domain similarity in adversarial robustness.

\clearpage
\bibliography{main}
\bibliographystyle{acl_natbib}

\clearpage
\appendix
\section{Appendix}

This appendix is organized as follows:
\begin{itemize}
    \item \textbf{Appendix \ref{ext_l2w}: } 
    We provide additional detail on our Learn2Weight architecture and Perturbation Sets Generation procedure. 
    \item \textbf{Appendix \ref{config}: } 
    We provide detail on the experimental configurations.
    \item \textbf{Appendix \ref{motivation}: } 
    We extend on similar-domain adversarial attacks by discussing the problem of joint distribution shifts.
    \item \textbf{Appendix \ref{theory1}-\ref{theory3}: } 
    We provide supporting theory that facilitated our development of the problem and defense method. 
\end{itemize}

\subsection{Learn2Weight (extended)}
\label{ext_l2w}

\noindent
\mybox{
\textbf{Learn2Weight \textnormal{(Algorithms \ref{alg:inf}, \ref{alg:train})}}
}
We first formalize meta learning architectures, in particular those adopting parameter adaptation with respect to distribution shift such as MAML \citep{finn2017modelagnostic} or hypernetworks \citep{ha2016hypernetworks}.
A 
meta learning system
$\mathscr{h}(\mathbf{x}, I) = \mathscr{f}(\mathbf{x}; \mathscr{mf}(\theta^{\mathscr{mf}}; I))$ 
is a pair of learners, the base learner $\mathscr{f}: \mathcal{X} \mapsto \mathcal{Y}$ and meta learner $\mathscr{mf}: \mathcal{I} \mapsto \Theta^{\mathscr{f}}$,
such that for the meta-data $I$ of input $\mathbf{x}$ (where $\mathcal{X} \mapsto \mathcal{I}$),
$\mathscr{mf}$ produces the base learner parameters $\theta_{I} = \mathscr{mf}(\theta^{\mathscr{mf}}; I)$.
The function $\mathscr{mf}(\theta^{\mathscr{mf}}; I)$ takes a conditioning input $I$ (e.g. meta-data, task header, few-shot samples, support set) to returns parameters $\theta_{I} \in \Theta^{\mathscr{f}}$ for $\mathscr{f}$. 
The meta learner parameters and base learner (of each respective distribution) parameters reside in their distinct parameter spaces $\theta^{\mathscr{mf}} \in \Theta^{\mathscr{mf}}$ and $\theta^{\mathscr{f}} \in \Theta^{\mathscr{f}}$.
The learner $\mathscr{f}$ takes an input $\mathbf{x}$ and returns an output $\hat{y} = \mathscr{f}(\theta_{I}; \mathbf{x})$ that depends on both $\mathbf{x}$ and the task-specific input $I$. 

In our implementation,
L2W is a sequence-to-sequence network \citep{10.5555/2969033.2969173},
not using any conditioning input $I$ (accepting $\mathbf{x}$ as sole input), 
computing the parameter change $\Delta \theta$ to an origin parameter $\theta_i$,
and only has access to its source distribution (no other domains, including the attacker's domain).

Our parameter adaptation architecture is implemented as follows.
At train-time,
\circled{1} $T$ training sets perturbed with respect to the source distribution are generated (Algorithm \ref{alg:tfo}). 
Their corresponding base learner parameters are computed.
The parameter differential is computed $\Delta \theta = \theta_j^{\mathscr{f}} - \theta_i^{\mathscr{f}}$
to obtain a meta-training set $\{\{ \mathbf{x}_j : \theta_j^{\mathscr{f}} - \theta_i^{\mathscr{f}} \}^{M}\}^{T}$.
\circled{2} $\mathscr{mf}$ (Algorithm \ref{alg:train}) optimizes its meta parameters $\theta^{\mathscr{mf}}$ containing the source and perturbed distributions, and their corresponding predicted weight differentials.
At test-time,
\circled{3} $\mathscr{mf}$ (Algorithm \ref{alg:inf}) computes the predicted weight differential to update the base learner parameters, and return a prediction $\hat{y} = \mathscr{f}(\theta_i+\mathscr{mf}(\mathbf{x}_j) ; \mathbf{x}_j)$.

\vspace{-0.5cm}
{\small
\begin{align}
\texttt{dist}(x_i, x_j) &= \sum_n^{N} \mathbf{P_i}(x_{n}, \mathcal{K}, N) \mapsto \Delta \theta \notag\\
\texttt{dist}(X_i, X_j) &= \sum_m^{M} \sum_n^{N} \mathbf{P_i}(x_{n, m}, \mathcal{K}, N) \mapsto \Delta \theta
\label{equation:diverg}
\end{align}}

\noindent
\mybox{
\textbf{Perturbation Sets \textnormal{(Algorithm \ref{alg:tfo})}}
}
Extending on Proposition 1, 
while Eqt 
\ref{equation:dataset_dist}
computed $\texttt{dist}(X_i, X_j)$ with respect to the distance between tokens in $\mathcal{K}$,
we can alternatively measure the distance in the total likelihood that each token exists in its $n$th index of a sequence $ \sum_n^{N} \mathbf{P_i}$ (Eqt \ref{equation:diverg}).

This motivates our construction of a perturbation set for use in constructing robust models.
A \textbf{perturbation set} is a set containing subsets of perturbed inputs. These perturbations may be generated with respect to 
word substitutions \citep{alzantot-etal-2018-generating, jia-etal-2019-certified},
Wasserstein balls \citep{DBLP:conf/icml/WongSK19},
or distribution shifts \citep{sinha2018certifiable, Sagawa2020Distributionally}.
The optimal $\theta^{\mathscr{mf}}$
is required to adapt $\theta^{\mathscr{f}}$ across
varying $ \sum_n^{N} \mathbf{P_i}$.
Hence, we are motivated to generate $T$ perturbation sets of diverging $ \sum_n^{N} \mathbf{P_i}$.
An average $\theta^{\mathscr{f}}$ (computed with static-adaptation methods) may not return high-accuracy across every point in the distribution of $ \sum_n^{N} \mathbf{P_i}$, further motivating the need for dynamic adaptation.
We pursue the following implementation (Algorithm \ref{alg:tfo}) to
(i) generate $T$ training sets of diverging $ \sum_n^{N} \mathbf{P_i}$ (representing varying ambiguities) for L2W to adapt $\theta^{\mathscr{f}}$ towards; 
(ii) avoid averaging randomly-generated $ \sum_n^{N} \mathbf{P_i}$ to a single static parameter;
(iii) sample shifted distributions that retain sufficient similarities to the source distribution (high dissimilarity returns random labels \textit{(Thm 1)}).

To construct one perturbation set (Eqt \ref{equation:perturbation_set}), we utilize an iterative minimax algorithm, 
where we iteratively apply a maximizing adversarial perturbation factor $\varepsilon \geq \varepsilon_{\textnormal{min}}$, 
and accept the batch of perturbed inputs if it yields a minimizing input distance $\texttt{dist} \leq d_{\textnormal{max}}$.
We repeat this $T$ times. 
To retain the relational property in Eqt \ref{equation:param_dist},
an optimal distance metric 
would be transfer loss. 
In-line with the rest of the paper, we retain FGSM as the adversarial attack algorithm.
The procedure for iterative perturbations are in-line with BIM \citep{https://doi.org/10.48550/arxiv.1607.02533}.
Iteratively evaluating perturbations to approximately invert a distance function $tf$
is in-line with \citet{10.1109/MILCOM.2016.7795300}'s inversion of $\frac{\partial \mathscr{f}(\theta; \mathbf{x})}{\partial \mathbf{x}}$.

\subsection{Experimental Configurations}
\label{config}

\textbf{Training}
We use a $80-20\%$ train-test split for both source data and domain-shifted data.
For LSTM (GRU) of 64 cells, 
tokens embedded w.r.t. GloVe,
$\texttt{sigmoid}$ ($\texttt{tanh}$) activation function,
randomly-initialized and trained with Adam optimizer and $80\%$ ($60\%$) dropout until early-stopping pauses training at loss 0.5.
For CNN accepting tokens embedded w.r.t. GloVe, of
3 convolutional layers with kernel widths of 3, 4, and 5, all with 100 output channels,
randomly-initialized and trained until early-stopping pauses training at loss 0.5.
We configure Logistic Regression based on \citet{Maas:2011:LWV:2002472.2002491} and default scikit-learn settings.
We initialize a pretrained BERT from huggingface (base, uncased) with its own embeddings
and trained until early-stopping pauses training at loss 0.5.
Practically,
base learner capacity is kept as low as possible to minimize the number of parameters needed to classify clean inputs, 
to therefore lower the capacity of a meta learner. 

\textbf{Defensive distillation}
In-line with \citet{papernot2016distillation, 10.1145/3052973.3053009}, we first train an initial model against target domain inputs and labels, and retrieve the raw class probability scores. 
The predicted probability values would be used as the new labels for the same target sentences, and we would train a new model based on this new label-sentence pair.

\textbf{Adversarial Training w.r.t. FGSM \& Perturbations Set}
In-line with \citet{goodfellow2014explaining,madry2017towards},
at each iteration we generate adversarial perturbations of $\varepsilon=0.4$ with FGSM, where the size of the perturbed set is the size of the batch.
For the Perturbation Sets variant,
we do not generate perturbations per batch/epoch,
and instead train the model on the pre-generated Perturbation Sets from Algorithm \ref{alg:tfo}.

\textbf{Fast Gradient Sign Method (FGSM)}
In-line with the FGSM implementation \citep{goodfellow2014explaining}, we generate adversarial samples by computing the sign of the gradient of the loss function w.r.t. inputs:

{\small $$\texttt{Adv}(\theta; \mathbf{x}; \varepsilon): \mathbf{x}^{\textnormal{adv}} = \mathbf{x} + \varepsilon \cdot \texttt{sign}(\frac{\partial \mathcal{L}(\theta; \mathbf{x}, y)}{\partial \mathbf{x}})$$}

Image inputs are continuous, while text inputs are discrete, hence the two considerations are (i) which tokens in a sequence to perturb and (ii) measuring the perturbation per token. 
In-line with \citet{10.1109/MILCOM.2016.7795300}, 
we iteratively insert perturbations until $f(\mathbf{x}^{\textnormal{adv}}) \neq y$.
Adapted from \citet{10.1109/MILCOM.2016.7795300}, 
the proportion of the sequence to be perturbed is $\varepsilon$ (and we randomly sample indices $n$ until $f(\mathbf{x}^{\textnormal{adv}}) \neq y$),
the perturbation measurement (sign of gradient direction) is based on FGSM, 
and the perturbed token $w$ is the closest token in $\mathcal{K}$ to the $\mathbf{x} + \varepsilon \cdot \texttt{sign}$
(where $\mathbf{x}$ would be interpreted as the position in the dictionary or embedding space).
We set $\varepsilon = 0.4$ for $\varepsilon \in [0, 1]$. 

\begin{lstlisting}[language=Python, mathescape=true]
while $\mathscr{f}(\theta; \mathbf{x}^{\textnormal{adv}}) \equiv y$:
    for $n \in I$ where $n \in [1, N]$, $|I| \leq \varepsilon N$:
        $w := x_n + \varepsilon \cdot \texttt{sign}(\frac{\partial \mathcal{L}(\theta; \mathbf{x}, y)}{\partial \mathbf{x}})$
        $\mathbf{x}^{\textnormal{adv}}_i \leftarrow w$
return  $\mathbf{x}^{\textnormal{adv}}$
\end{lstlisting}    
    
\textbf{Learn2Weight}
We trained a sequence-to-sequence network \citep{10.5555/2969033.2969173} on each sequence pair $\mathbf{x} : \Delta \theta$.
$\Delta \theta$ is a flattened $1 \times |\theta|$-dimensional vector, hence the input sequence length is $N$ and output sequence length is $|\theta|$.
We enlarge the capacity of the sequence-to-sequence network $\mathcal{C}^{\mathscr{mf}}$ with respect to the capacity (parameter count) of the base learner such that the loss of the meta learner converges $\mathcal{C}^{\mathscr{mf}} \geq \mathcal{C}^{\mathscr{f}}$ \textit{s.t.} $\mathcal{L}^{\mathscr{mf}}(\mathbf{x}, \Delta \theta)$,
similar to \citet{vonoswald2020continual} and \citet{52ceba2916624178952b02a128b8824e} 
(we do not have a clear-cut ratio, as different architectural differences affect the compression ratio; for example, a wider but shallower base learner, with the same number of parameters as a narrower but deeper base learner, can be learnt by a smaller meta learner).
This results in fine-tuned changes in sequence length, units/cells, etc.
The meta learner is trained until early-stopping pauses at loss 0.5.

With respect to the constraints and formulation in Hypothesis 2, 
there are some distinct differences between L2W and other meta learning architectures such as MAML \citep{finn2017modelagnostic} or hypernetworks \citep{ha2016hypernetworks}.
Namely, 
we do not use any conditioning input $I$ (accepting $\mathbf{x}$ as sole input), 
and we compute the parameter change $\Delta \theta$ to an origin parameter $\theta_i$ (i.e. the parameter space for different distributions are all identical $\Theta_i \equiv \Theta_j$, given Proposition 2).
For our implementation, 
$\texttt{dist}(\theta_i, \theta_j) \equiv \Delta \theta$.
Additional problem setup constraints are that
the input space (with respect to the word/character embedding space in language/text) is bounded, 
the model only has access to its own source distribution (no other data from other domains/distributions),
and the label set is finite and small.

\textbf{Perturbation Sets}
Our configurations are as follows:
10 perturbation sets, 10 maximum iterations for generating perturbations, transfer loss distance metric, maximum distance for transfer loss of 0.1, initial $\varepsilon = 0.9$ with perturbation learning rate of 0.05.
We subsequently provide further details on our supporting motivation and theory for the use of perturbation sets.

Rather than optimizing $\theta^{\mathscr{mf}}$ towards adapting $\theta^{\mathscr{f}}$ with respect to adversarial tokens $x^{\textnormal{adv}}_n$ (which represent perturbation-specific adaptation that may not arise at test-time), 
we aim to optimize $\theta^{\mathscr{mf}}$ towards adapting $\theta^{\mathscr{f}}$ with respect to varying $ \sum_n^{N} \mathbf{P_i}$ (which represent structure-specific adaptation, where for example an increase in ambiguous structural patterns or unknown tokens should correspondingly result in sparser feature selection in adapted parameters).
In particular, we would like to generate $T$ training sets that include diverging $ \sum_n^{N} \mathbf{P_i}$ such that 
{\small 
$\{ \texttt{dist}(\sum_m^{M} \sum_n^{N} \mathbf{P_i}(x_{i, n, m}), \sum_m^{M} \sum_n^{N} \mathbf{P_i}(x_{t, n, m}) ) \}^{t \in T}$
},
to train L2W to adapt $\theta^{\mathscr{f}}$ to varying ambiguities represented by $ \sum_n^{N} \mathbf{P_i}$.

The use of perturbation sets, adversarial samples, and augmented data samples during training do indeed generate diverging $ \sum_n^{N} \mathbf{P_i}$. 
However, these static-adaptation methods map an average diverging $ \sum_n^{N} \mathbf{P_i}$ to a single (robust) parameter: 
$\frac{1}{M} \sum_m^{M} \sum_n^{N} \mathbf{P_i}(x_{n, m}, \mathcal{K}, N) \mapsto \Delta \theta $.
While multiple perturbation sets within the total $T$ sets may have similar $ \sum_n^{N} \mathbf{P_i}$,
the required parameter adaptation $\Delta \theta$ for each unique $ \sum_n^{N} \mathbf{P_i}$
may differ, and thus benefit from a meta learner performing parameter-switching/adaptation.
This is empirically validated in Table 3,
where static-adaptation methods (even on our shifted perturbations set) perform sub-optimally across all domain pairs.

The perturbation sets we construct are based solely on the source distribution, but should contain perturbations that resemble that originating from a joint distribution shift. 
We would like to sample distributions that have a high proportion of perturbations such that they are distant in the input space $\uparrow \texttt{dist}(X_i, X_j)$.
It is observed from \citet{zhang2017understanding} that if the perturbed distributions are too dissimilar such that they tend to be random sentences or labels are randomly flipped, a base learner will be able to overfit and memorize these perturbed training instances, but will retain no generalizability at test-time. 
Hence, at the same time we would like to sample distributions to retain sufficient similarities to the source distribution, as indicated by Table 2 that natural domains have high SharedVocab and low transfer loss with respect to a source distribution. 
Distributions of high input distances are not expected at test-time (evaluation of hypothesis 1 indicates high-distance distributions would yield random predictions, and similar domains yield lower accuracy than dissimilar domains).
Bounding the maximum distance in the input space also assists in bounding the parameter space,
increasing the ease for L2W to generalize its output (base learner parameter) space.

An important objective is to ensure the transfer loss of the perturbation set is small/bounded and diverse,
but we also regulate the $\varepsilon$, in addition to the aforementioned reasons, for the practical reason that not regulating $\varepsilon$ may result in the perturbation set containing no perturbations ($\varepsilon = 0$) to attain a low $tf$ (i.e. $X_i \equiv X_j$), thus to avoid this we ensure perturbations must manifest at a mimimum extent $\varepsilon_{\textnormal{min}}$.

\subsection{Joint Distribution Shifts: Beyond Similar-domain Adversarial Attacks}
\label{motivation}

\noindent\textbf{Motivation. }
While our work thus far empirically validates the specific combination of label shift (adversarial attack) and domain shift, 
it highlights a deeper concern on robustness against joint distribution shifts. 
In the subsequent sections, we motivate theoretical grounds on robustness to joint distribution shifts in general, while also applicable to L2W and the similar-domain adversarial attack.

The problem of distribution shift, the divergence between the train-time and test-time distributions, manifests in different forms in NLP.
Robustness against adversarial attacks, a form of label shift, 
have resulted in defenses such as such as adversarial training \citep{goodfellow2014explaining,madry2017towards},
defensive distillation \citep{papernot2016distillation, 10.1145/3052973.3053009}, 
denoising \citep{DBLP:conf/cvpr/LiaoLDPH018},
many of which return parameters that are static at test-time. 
Robustness against domain shifts has resulted in various methods of adaptation from a source distribution to a target distribution \citep{pmlr-v97-liu19b, ziser-reichart-2019-task, 10.1007/978-3-319-71246-8_46}.

In particular, we are motivated to study the problem of joint distribution shifts, where a target distribution can manifest perturbations from multiple sources distributions simultaneously. 
Given the bounded input space and non-conformity of perturbation sources, 
we evaluate a parameter adaptation strategy based in meta learning, 
where adapted parameters are mapped for distant tasks/distributions in the input space. 
We construct a meta learner, Learn2Weight, 
trained on perturbation sets as an interpolating input-parameter adaptation model.

We validate these two hypothesis for robustness to joint distributional shift in NLP:
\begin{enumerate}[noitemsep,topsep=0pt,parsep=0pt,partopsep=0pt]
    \item
    \mybox{\textbf{Hypothesis 1}}
    Model performance weakens when a test-time input manifests joint distribution shift, evaluated on adversarial and domain-specific perturbations. The performance worsens when a similar domain is used.
    \item 
    \mybox{\textbf{Hypothesis 2}}
    A meta learner adapting the model parameters with respect to perturbed inputs can retain robust performance.
\end{enumerate}

We ground our work in the problem of joint distribution shift,
showing not only reduced performance to models in its exposure, 
but additional performance reduction when the domain shift is similar to the source distribution.
With perturbation set construction motivated by our understanding of the input and parameter space bounds,
and motivated by dynamic parameter adaptation,
our proposed adaptation strategy, Learn2Weight, tackles joint distribution shifts in NLP settings. 

\noindent\textbf{Related Works. }
There is sparsely-growing literature on joint distribution shifts, and we would be amongst the first in an NLP setting to study joint shifts and their methods for robustness.
\citet{https://doi.org/10.48550/arxiv.2205.09891}
construct a compressed parameter subspace to 
return optimal parameters for various test-time distribution shifts,
including label shift (backdoor, adversarial, permutation, rotation perturbations),
domain shift, and task shift.
\citet{qi2021mind}
used text style transfer to perform adversarial attacks.
\citet{naseer2019crossdomain}
generated domain-invariant adversarial perturbations to fool models of different domains.
\citet{10.5555/2946645.2946704} proposed  domain-adapted adversarial training to improve domain adaptation.
\citet{geirhos2018imagenettrained} demonstrated the use of stylized perturbations with AdaIN 
can improve performance on corruptions dataset ImageNet-C. 
AdvTrojan \citep{liu2021synergetic} combined adversarial perturbations with backdoor trigger perturbations to craft stealthy triggers to perform backdoor attacks.
\citet{santurkar2020breeds} synthesized distribution shifts by combining random noise, adversarial perturbations, and domain shifts
to contribute subpopulation shift benchmarks.
\citet{10.1007/978-3-030-58580-8_4} proposed a robustness measure by augmenting a dataset with both adversarial noise and stylized perturbations, by evaluating a set of perturbation types including Gaussian noise, stylization and adversarial perturbations.
\citet{datta2022backdoors} demonstrated the low likelihood of backdooring a model in the presence of joint distribution shifts, including multiple perturbations of the same shift type (backdoor), and multiple perturbations of different shift types (backdoor, adversarial, domain).
\citet{datta2022low} demonstrated a low-loss compressed subspace defense to tackle joint distribution shifts in a multi-agent backdoor attack setting.

\subsection{Propositions}
\label{theory1}

\noindent\mybox{\textbf{Proposition 1}}
\textit{
Distribution shifts, manifesting as perturbations (in this case, adversarial and domain-specific), in NLP are bounded within a finite dictionary or embedding space. 
Any shifted distribution $\mathcal{X}_j$ is located at a bounded distance from an (origin) source distribution $\mathcal{X}_i$.
}

A \textbf{distribution shift} is denoted as the divergence between a source (train-time) and target (test-time) distribution. 
The cause for shift can vary by \textit{source} of distribution (e.g. domain shift, task shift, label shift) and \textit{variations} per source (e.g. multiple backdoor triggers, multiple domains).
A \textbf{joint distribution shift} is denoted distribution shift attributed to multiple sources and/or variations per source; this is in contrast to \textbf{disjoint distribution shift}, attributed to a single source and variation of shift.

We denote a token $x$, which is an index mapped to a word/character (word, in our evaluation) in a finite and discrete dictionary (or embedding space) $\mathcal{K}$ where $|\mathcal{K}| = k$ (GloVe \citep{pennington2014glove}, in our evaluation).
A sentence is an $N$-token sequence $\mathbf{x} = \{x\}^{N}$.
A dataset containing $M$ labelled sequences is constructed as $X = \{ \mathbf{x} \}^M$ and mapped to their corresponding indexed labels $\mathbf{x} \mapsto y$, $X \mapsto Y$.

We denote $\mathcal{X}$ as the input space. In the NLP setting, as $\mathcal{K}$ is discrete and finite, the maximum distance between any 2 arbitrary tokens (diameter of $\mathcal{X}$) approximates the bound in our evaluation $d_{\textnormal{max}} := \max_{\mathbf{x_i}, \mathbf{x_j} \sim \mathcal{X}} \texttt{dist}(\mathbf{x_i}, \mathbf{x_j})$.
We denote a generic distance metric $\texttt{dist}$, and the properties of the metric should be inferred from the input arguments.
To retain generality, 
we refer to the input (label) distribution as its corresponding input (label) space.
A distribution shift $\mathcal{X}_i \rightarrow \mathcal{X}_j$ would be measured as the relative distance between the 2 subspaces $\texttt{dist}(\mathcal{X}_i, \mathcal{X}_j)$.
Any point in these subspaces reside in $\mathcal{X}$ where $\mathcal{X}_i, \mathcal{X}_j \subseteq \mathcal{X}$,
hence the distribution shift is also bounded $\texttt{dist}(\mathcal{X}_i, \mathcal{X}_j) \leq d_{\textnormal{max}}$.
A perturbation $\varepsilon_n$ is a change in value of the token $x$ w.r.t. $\mathcal{K}$ at the $n$th position of the sequence such that $\mathbf{x_j} = \{ x_{i} \}^{n \notin N} \cup \{ x_{i, n} + \varepsilon_n \}^{n \in N}$.
We denote $\varepsilon$ as a vector composed of elements from set $\{ 0, \varepsilon_n \}$ such that $\mathbf{x_j} = \mathbf{x_i} + \varepsilon$, 
where the position $n$ is computed by $\texttt{Adv}$.
Perturbations can manifest as adversarial perturbations and/or domain-specific perturbations.
As perturbations can cause an input point to shift from one subspace to another subspace, we compose distribution shifts in terms of perturbations, and transitive to shifts, perturbations are also bounded within $\mathcal{K}$ and $\mathcal{X}$. 

Suppose $f_i$ is the ground-truth function that defines the source distribution (subspace) of index $i$ mapping to the label distribution $\mathcal{X}_i \mapsto \mathcal{Y}_i$ such that $y = f_i(\{x\}^{N}_{x \sim \mathbf{P_i}(\mathcal{K}, N)})$, where each token $x$ is sampled from $\mathcal{K}$ to form a sequence of length $N$ and returns ground-truth label $y$.
$\mathscr{f}$ is a function approximating $f$, accepting the input sequence $\mathbf{x}$ and model parameters (approximating $f_i$) $\theta_i$ as arguments to return the predicted label $\hat{y} = \mathscr{f}(\theta_i; \mathbf{x})$.
$\mathscr{f}$ is a generalized function, specifically a deep neural network of fixed architecture with respect to varying $\mathbf{x}$, and adaptation to $\mathscr{f}$ with respect to varying $f_i$ is through varying $\theta_i$.
$\theta_i$ is obtained through Stochastic Gradient Descent by minimizing the loss between the actual and approximated functions through the enumeration of samples from a dataset $\{x\}^N \in X_i$ such that
$\theta_{i, t} := \theta_{i, t-1} - \sum_{\{x\}^N, y}^{X, Y} \frac{\partial \mathcal{L}(\{x\}^N, y)}{\partial \theta} $.
As validated in \citet{galanti2020modularity}, we retain the assumption that a mapped $\mathscr{f}(\theta; X)$ exists for any sampled distribution $X \sim \mathcal{X}$. 

Moreover, $f_i \mapsto \mathbf{P_i}(\mathcal{K}, N)$ the function is mapped to its corresponding probability density function, which is used for sampling tokens to construct a sequence from distribution $\mathcal{X}_i \mapsto \mathbf{P_i}(\mathcal{K}, N)$. 
$\mathbf{P_i}(\mathcal{K}, N)$ maps each item in $\mathcal{K}$ to the probability of occurrence of token $x_n$ at the $n$th position ($n \in N$) of a sequence in source distribution $\mathcal{X}_i$, i.e. $\{\mathcal{K}, N\} \mapsto \mathbf{P_i}(\mathcal{K}, N)$.
This could also be approximated as obtaining the hidden state representations or activations returned at the $\ell$th layer of a model $\mathscr{f}(\cdot; \theta_i)$ (e.g. BERT \citep{devlin-etal-2019-bert}): $\mathbf{P_i} \approx h_{\ell} = \mathscr{f}_{\ell}(\theta_i; \cdot)$.

Distribution shift can manifest as domain shift $\mathcal{X}_i \rightarrow \mathcal{X}_j$, which tend to manifest as a change in the underlying source distribution.
Domain shift and/or text distance between 2 datasets $\texttt{dist}(X_i, X_j)$ 
could be measured by the 
Kullback-Leibler divergence, 
Maximum Mean Discrepancy \citep{10.1093/bioinformatics/btl242}, 
Word Mover’s Distance \citep{pmlr-v37-kusnerb15}, 
Transfer Loss \citep{blitzer2007biographies,Glorot:2011:DAL:3104482.3104547}, etc. 
As each input sequence is a $1 \times N$-dimensional vector, we can compute the distance between 2 datasets w.r.t. the average distance of their contained sequences, using any of the aforementioned distance metrics: 

{\small
\begin{align}
\texttt{dist}(X_i, X_j)
& = \mathbb{E}_{\mathbf{x}_i, \mathbf{x}_j \sim X_i, X_j}[\texttt{dist}(\mathbf{x}_i, \mathbf{x}_j)] 
\label{equation:dataset_dist}
\\
& \approx \mathbb{E}_{\mathbf{x}_i, \mathbf{x}_j \sim X_i, X_j}||\mathbf{x}_i, \mathbf{x}_j||_2^2 \notag
\end{align}}

Unlike domain shift, adversarial perturbations tend to manifest as a change in the label distribution, specifically a change in the mapping between the source and label distribution). 
Perturbations from adversarial attack algorithms, due to the discrete nature of text input spaces,
tend to manifest as 
token substitutions (i.e. a change to the token sampling strategy to construct an input sequence $\mathbf{x}$).
Thus, we can generalize the construction of an adversarially-perturbed set as: 

\begin{equation}
\small
\begin{split}
X^{\textnormal{adv}}_i 
& = \{ \{x\}^{N, \textnormal{adv}}_{x \sim \mathbf{P_i}(\mathcal{K}, N)} \}^{M} \\
& := \max \sum_m^M |y - f_i(\{x\}^{N}_{x \sim \mathbf{P_i}(\mathcal{K}, N)})| \\
& := \max \sum_m^M |y - \mathscr{f}(\theta_i; \{x\}^{N}_{x \sim \mathbf{P_i}(\mathcal{K}, N)})|
\end{split}
\end{equation}

\noindent\mybox{\textbf{Proposition 2}}
\textit{A distribution shift in the datasets $X_i \rightarrow X_j$ in a bounded input space can be approximated by the distance between their optimized parameters $\theta_i \rightarrow \theta_j$ and difference in optimization trajectories per epoch. }

Suppose we train models optimized for different shifted distributions starting from a constant, shared initialization (e.g. 1 random $\theta$, 1 pre-trained $\theta$).
For 2 independent parameter optimization processes of distributions $\mathcal{X}_i$ and $\mathcal{X}_j$, 
in order for $\theta_i$ and $\theta_j$ to reside in a local subspace in the parameter space (i.e. minimize distance between parameters 
$\texttt{dist}(\theta_{i}, \theta_{j}) \approx 0$
), the difference in gradient updates across their training epochs (assuming the same total epoch count $\mathbf{E}$) should be minimal, where
\begin{equation}
\small
\begin{split}
\texttt{dist}(\theta_{i}, \theta_{j})
=
\sum_e^{\mathbf{E}} \Bigg| \frac{\partial \mathcal{L}(X_i, Y_i)}{\partial \theta_{i, e}} - \frac{\partial \mathcal{L}(X_j, Y_j)}{\partial \theta_{j, e}} \Bigg|
\end{split}
\label{equation:param_dist}
\end{equation}

Measurements for $\texttt{dist}(\theta_{i}, \theta_{j})$ 
include cosine distance \citep{wortsman2021learning},
centered kernel alignment \citep{pmlr-v97-kornblith19a},
set difference in subnetworks \citep{datta2022low}.
We formulate the parameter distance as the distance in successive gradient updates based on a set of observations (not intended to be mutually-exclusive). 
$E \in [0, \mathbf{E}]$ is an arbitrary epoch.

\textbf{(Observation 1)}
\textit{ $\theta_i$ and $\theta_j$ converge (diverge) if their training distributions contain transferable (interfering) features.}

The transfer-interference trade-off \citep{riemer2019learning} finds that if 2 arbitrary inputs for 2 independently trained networks contain transferable features, the gradient updates share the same direction; the updates share opposite directions if features interfere against each other. 
Specifically, we denote \textit{transfer} and \textit{interference} as:
\begin{equation}
\small
\begin{cases} \textnormal{Transfer: } \frac{\partial \mathcal{L}(X_i, Y_i)}{\partial \theta} \cdot \frac{\partial \mathcal{L}(X_j, Y_j)}{\partial \theta} > 0 \\ 
\textnormal{Interference: } \frac{\partial \mathcal{L}(X_i, Y_i)}{\partial \theta} \cdot \frac{\partial \mathcal{L}(X_j, Y_j)}{\partial \theta} < 0 \end{cases}
\nonumber
\end{equation}
\citet{NEURIPS2020_0607f4c7} finds that 2 parameters trained on 2 different distributions (e.g. domain-shifted)
initialized on pre-trained weights
will be optimized towards a shared flat basin in the loss landscape (though the likelihood for this occurrence weakens when a constant random initialization, or 2 independent random initializations are used).
This body of work supports the notion that for 2 distributions $\mathcal{X}_i$ and $\mathcal{X}_j$,
if the 2 distributions contain transferable features, then $|\texttt{dist}(\theta_{i}, \theta_{j})| \approx 0$ for $e \leq E$.
If the 2 distributions contain interfering features, then $|\texttt{dist}(\theta_{i}, \theta_{j})| > 0$ for $e > E$.
As the 2 distributions diverge in similarity, we would expect $E$ to decrease so that the parameters can diverge accordingly $\texttt{dist}(X_{i}, X_{j}) \propto \frac{1}{E}$; i.e. if the 2 distributions are the same or similar, $E \approx \mathbf{E}$.

\textbf{(Observation 2)}
\textit{ $\theta_i$ and $\theta_j$ may diverge, attributing to the presence of highly-contextual/semantic features (to optimize for feature density), or robust features (to optimize for feature sparsity).}

Optimizing model parameters along the density-sparsity trade-off
is a needed consideration.
Parameters optimized for feature \textit{density} pertains towards features that are more input instance-specific or spuriously-correlated (e.g. features that are highly-specific to a given instance's distribution, context, or semantics).
Parameters optimized for feature \textit{sparsity} pertains towards features that are more task-specific (e.g. features that are relatively agnostic to instance-specific features, but highly-specific to a given task's distribution, context, or semantics).
For example, a feature dense model 
would classify between a dolphin and a dog
based on non-object-specific features, such as whether the background is blue.
Robustness training procedures (e.g. data augmentation \citep{yun2019cutmix}, adversarial training \citep{goodfellow2014explaining,madry2017towards}) and sparse model training procedures (e.g. model pruning, dropout \citep{JMLR:v15:srivastava14a}, contrastive learning \citep{pmlr-v139-wen21c} demonstrate improved model robustness and/or generalizability through sparse feature selection. 
These work show that robust (sparse; more distributionally-robust) features can be learnt to improve performance against natural or synthetic perturbations.
In addition, many of these methods are static-adaptation techniques, where different methods and different hyperparameters per method return varying robustness accuracies (i.e. different model parameters). This indicates there is no one-fit-all parameter optimization strategy against different types of perturbations. 
This is shown in \citet{https://doi.org/10.48550/arxiv.2203.05482}, where the authors construct a "soup" of models with varying augmentation techniques and hyperparameters to maximize robustness accuracy against varying types of perturbations.
For example, the end-goal is not necessarily that we should optimize parameters for maximum feature sparsity or maximum feature density. 

Parameter adaptation may also fall outside of the aforementioned observations. \citet{Rame_2021_ICCV} and \citet{havasi2021training} showed that a wide network can learn multiple subnetworks that may operate independently from each other.

From Observations 1 and 2, we aim to clarify that if we shift a source distribution $\mathcal{X}_i \rightarrow \mathcal{X}_j$, it does not equate to the generation of interfering features, or the inducement of SGD to optimize towards feature density/sparsity. 
Different perturbations or distances between distributions in the input space may result in different parameter optimization strategies; for example, increasing adversarial perturbation factor $\varepsilon$ or increasing domain dissimilarity $\texttt{dist}(X_i, X_j)$ may not equate to a linear increase in feature sparsity of $\theta$. 
One of the implications of this observation is that, 
though $X_i$ and $X_j$ may contain transferable features (due to similarity), if $\theta_i$ and $\theta_j$ are optimized independently, and either or both $X_i$ and $X_j$ contain features that induce density/sparsity, then it would result in divergence between $\theta_i$ and $\theta_j$ early in training: {\small $|\texttt{dist}(\theta_{i}, \theta_{j})| > 0$ for $\begin{cases}  e \leq E\\ e > E\end{cases}$}.

In particular, we note that a linear change in adversarial perturbation factor $\varepsilon$ or domain similarity may not translate into a linear change in distance in the input space $\texttt{dist}(X_i, X_j)$. 
For example, if a low $\varepsilon$ breaks the semantic structure of a sentence, then $\texttt{dist}(X_i, X_j)$ may be perceived to be high. 
Based on this discussion, 
we approximate the distance between 2 distributions in a bounded input space by the distance between their optimized parameters trained on a constant initialization (Eqt \ref{equation:param_dist}); as no corresponding ground-truth input distribution may exist for the constant random initialization, the mapped origins for the input space $\mathcal{X}$ and parameter space $\Theta$ are the source distribution $\mathcal{X}_i$ and its mapped parameters $\theta_i$ respectively.

From Observations 1 and 2,
we intend to make it clear that parameter adaptation
does not mean 
a linear change in $\texttt{dist}(\mathcal{X}_i, \mathcal{X}_j)$ will map to a specific pattern of parameter change (e.g. feature sparsity/density, transfer/interference).
Conversely, 
a linear change in adversarial perturbation factor $\varepsilon$ and/or domain similarity may not map to a linear change in distance in the input space $\texttt{dist}(\mathcal{X}_i, \mathcal{X}_j)$. 
As the mapping between the input-output spaces $\mathcal{X} \rightarrow \Theta$ may not be linearly-interpolatable, 
the adaptation function should be a non-linear function.
As the practical objective is to compute $\texttt{dist}(\theta_i, \theta_j)$
in order to compute $y$,
we can inversely measure 
$\texttt{dist}(X_{i}, X_{j})$ w.r.t. $\texttt{dist}(\theta_i, \theta_j)$
to simplify further analysis.
We approximate the distance between 2 distributions in a bounded input space by the distance between their optimized parameters trained on a constant initialization.
As no corresponding ground-truth input distribution may exist for the constant random initialization, the mapped origins for the input space $\mathcal{X}$ and parameter space $\Theta$ are the source distribution $\mathcal{X}_i$ and its mapped parameters $\theta_i$ respectively.

{\small
\begin{align}
\texttt{dist}(X_{i}, X_{j}) 
& \propto
\texttt{dist}(\theta_{i}, \theta_{j}) \\ 
& \propto E \Bigg| \sum_e^{E} \bigg| \frac{\partial \mathcal{L}(X_i, Y_i)}{\partial \theta_{i, e}} - \frac{\partial \mathcal{L}(X_j, Y_j)}{\partial \theta_{j, e}} \bigg| \notag
\label{equation:param_dist}
\end{align}}

\noindent\mybox{\textbf{Proposition 3}}
\textit{
Suppose 
the $(\ell-1)$th layer in an $\ell$-layer neural network $\mathscr{f}$ is the layer returning class probabilities such that $\mathscr{f}_{\ell-1}(\theta; \mathbf{x}) = \{y: \rho\}$.
and perturbations per sequence $\xi = \sum_{\lambda}^{\Lambda} \varepsilon_{\lambda}$, where $\Lambda$ are different sources/variations of shift.
To mitigate the increase in error attributed to $\argmax_{y \sim \mathcal{Y}} \mathscr{f}_{\ell-1}(\theta_i; \xi)$, 
for a shifted input $\mathbf{x}_j$,
we may adapt the parameter $\theta_i \rightarrow \theta_j$ to converge class probabilities w.r.t. $\xi$ to 0,
where 
{\small
$\begin{cases}
\mathscr{f}_{\ell-1}(\theta_i; \xi) = \sum_{\lambda}^{\Lambda} \{y_{\varepsilon_{\lambda}}: \rho_{\varepsilon_{\lambda}}\} \\
\mathscr{f}_{\ell-1}(\theta_j; \xi) = \sum_{\lambda}^{\Lambda} \{y_{\varepsilon_{\lambda}}: 0\}
\end{cases}$
}.
}

By distance metric (Eqt \ref{equation:dataset_dist}), we can decompose the distribution shift between 2 datasets into a set of perturbations per sequence $\xi = \sum_{\lambda}^{\Lambda} \varepsilon_{\lambda}$, where $\Lambda$ are different sources/variations of shift (e.g. domains, adversarial perturbation). 

\begin{equation}
\small
\begin{split}
X_j - X_i
& = \{\{x_j\}^N\}^M - \{\{x_i\}^N\}^M \\
& = \{\{\xi\}^N\}^M  \\
& = \{\{\sum_{\lambda}^{\Lambda} \varepsilon_{\lambda}\}^N\}^M 
\end{split}
\end{equation}

For an $\ell$-layer neural network $\mathscr{f}$, 
suppose the $(\ell-1)$th layer is the layer before the prediction layer that returns class probabilities $\{y: \rho\}$ such that $\mathscr{f}_{\ell-1}(\theta; \mathbf{x}) = \{y: \rho\}$.
This results in a decomposition of the class probabilities altered with respect to $\xi$:

\begin{equation}
\small
\begin{split}
\mathscr{f}_{\ell-1}(\theta_i; \mathbf{x}_j)
& = 
\argmax_{y \sim \mathcal{Y}} \{ \mathscr{f}_{\ell-1}(\theta_i; \mathbf{x}_i) + \sum_{\lambda}^{\Lambda} \mathscr{f}_{\ell-1}(\theta_i; \varepsilon_{\lambda}) \} \\
& =
\argmax_{y \sim \mathcal{Y}} \{ 
\{y_{\mathbf{x}_i}: \rho_{\mathbf{x}_i}\} + \sum_{\lambda}^{\Lambda} \{y_{\varepsilon_{\lambda}}: \rho_{\varepsilon_{\lambda}}\}
\} 
\end{split}
\end{equation}

In particular, we find that distribution shift results in a misclassication (reduction in accuracy) when the perturbations $\xi$ shifts the class probabilities towards a different class. 

\begin{equation}
\small
\begin{split}
& \mathscr{f}_{\ell-1}(\theta_i; \mathbf{x}_j) - \mathscr{f}_{\ell-1}(\theta_i; \mathbf{x}_i) \\
& =
\argmax_{y \sim \mathcal{Y}} \{ \{y_{\mathbf{x}_i}: \rho_{\mathbf{x}_i}\}
+ \sum_{\lambda}^{\Lambda} \{y_{\varepsilon_{\lambda}}: \rho_{\varepsilon_{\lambda}} \} \}  
\\ & \phantom{=} 
- 
\argmax_{y \sim \mathcal{Y}} \{y_{\mathbf{x}_i}: \rho_{\mathbf{x}_i}\}
\\
& = 
\begin{cases}
\argmax_{y \sim \mathcal{Y}} \{ 
\sum_{\lambda}^{\Lambda} \{y_{\varepsilon_{\lambda}}: \rho_{\varepsilon_{\lambda}}\} 
\} \\
\phantom{=========} \textnormal{\textit{if}} \argmax_{y \sim \mathcal{Y}} \{ \sum_{\lambda}^{\Lambda} \{y_{\varepsilon_{\lambda}}: \rho_{\varepsilon_{\lambda}}\} \} \\
\phantom{=} \equiv 
\argmax_{y \sim \mathcal{Y}} \{ \{y_{\mathbf{x}_i}: \rho_{\mathbf{x}_i}\}
+ \sum_{\lambda}^{\Lambda} \{y_{\varepsilon_{\lambda}}: \rho_{\varepsilon_{\lambda}} \} \} 
\\
0 \phantom{========} \textnormal{\textit{otherwise}}
\end{cases} 
\\
& = 
\begin{cases}
\argmax_{y \sim \mathcal{Y}} \mathscr{f}_{\ell-1}(\theta_i; \xi) \\
\phantom{=========} \textnormal{\textit{if}} \argmax_{y \sim \mathcal{Y}} \{ \sum_{\lambda}^{\Lambda} \{y_{\varepsilon_{\lambda}}: \rho_{\varepsilon_{\lambda}}\} \} \\
\phantom{=} \equiv 
\argmax_{y \sim \mathcal{Y}} \{ \{y_{\mathbf{x}_i}: \rho_{\mathbf{x}_i}\}
+ \sum_{\lambda}^{\Lambda} \{y_{\varepsilon_{\lambda}}: \rho_{\varepsilon_{\lambda}} \} \} 
\\
0 \phantom{========} \textnormal{\textit{otherwise}}
\end{cases}
\end{split}
\label{equation:joint_shift}
\end{equation}

To mitigate the increase in error attributed to $\argmax_{y \sim \mathcal{Y}} \mathscr{f}_{\ell-1}(\theta_i; \xi)$, 
for a shifted input $\mathbf{x}_j$,
we would need a corresponding adapted parameter $\theta_i \rightarrow \theta_j$,
where 
{\small
$\begin{cases}
\mathscr{f}_{\ell-1}(\theta_i; \xi) = \sum_{\lambda}^{\Lambda} \{y_{\varepsilon_{\lambda}}: \rho_{\varepsilon_{\lambda}}\} \\
\mathscr{f}_{\ell-1}(\theta_j; \xi) = \sum_{\lambda}^{\Lambda} \{y_{\varepsilon_{\lambda}}: 0\}
\end{cases}$
}.
\newline
$\theta_i$ is the non-optimal parameters and perturbations can shift the class probabilities;
$\theta_j$ is the optimal, adapted parameters and converge the class probabilities w.r.t. $\xi$ to 0.

\subsection{Hypotheses}
\label{theory2}

Extending on Proposition 3,
the objective of this work is \textit{to propose an adaptation technique against joint distribution shifts}. 
We evaluate Hypotheses 1 and 2, the former proposing a suitable joint distribution setting, the latter proposing an adaptation method to tackle the scenario in Hypothesis 1. 
The hypotheses are affirmative: their intention is to illustrate the theoretical grounding for which the subsequent empirical observations will validate.

\noindent\mybox{\textbf{Hypothesis 1}}
\textit{It is hypothesized that a perturbed sample manifesting joint distribution shift, in particular adversarial shift and domain shift (worse, similar domain shift), the model's accuracy w.r.t. the perturbed input would be lower.}

\textbf{(Hypothesis 1a)}
\textit{It is hypothesized that an input manifesting joint distribution shift will yield lower model accuracy on shifted inputs.}
At train-time, we presume a model is trained on a source dataset $X_i \mapsto Y_i$.
At test-time, we presume a model encounters a sample that is joint-distributionally-shifted from $X_i$; we adopt 2 sources of perturbations $\xi$, (i) adversarial perturbations $\varepsilon_{\textnormal{adv}}$, and (ii) domain shift $\varepsilon_{\textnormal{domain}}$ from dataset $X_j \mapsto Y_j$.
Proposition 1 implies that despite the insertion of perturbations, the shifted distribution is a bounded distance from the source training distribution. 
However, Proposition 2 also indicates that trajectory-changing perturbations, such as the replacement of semantic/contextual structure through domain shift, will result in different optimal model parameters, implying an increased distance between the 2 distributions in the input space.
If trajectory-changing perturbations manifest, then passing $X_j$ through a model with $\theta_i$ will likely result in the label-shifting case in Eqt \ref{equation:joint_shift}, and thus reduce model accuracy.

\textbf{(Hypothesis 1b)}
\textit{It is hypothesized that an adversarially-perturbed input originating from a similar domain to the training domain will return a lower accuracy on perturbed inputs than that of a dissimilar domain, for a small number of labels in $\mathcal{Y}$ such that $\frac{1}{|\mathcal{Y}|} \nrightarrow 0$.}
Extending on Hypothesis 1a, 
we further hypothesize that adversarial inputs generated from a similar domain to the training distribution will result in a lower accuracy w.r.t. perturbed inputs than a dissimilar domain. 
Intuitively, one would expect a dissimilar domain to return a lower accuracy w.r.t. perturbed inputs, given the expectedly larger change in semantic structure, i.e. $\texttt{dist}(X_{i}, X_{j}) \uparrow \propto \texttt{dist}(\theta_{i}, \theta_{j}) \uparrow$.
This hypothesis would be supported by literature in out-of-distribution robustness \citep{https://doi.org/10.48550/arxiv.1610.02136}.
Based on Proposition 2, 
$\theta_{i}$ and $\theta_{j}$ may be more distant w.r.t. dissimilar domains than similar domains. 
Though $\theta_{i}$ and $\theta_{j}$ are distant, the distance between parameters do not equate to reduction in accuracy. 
Theorem 1 shows
that when a test-time distribution is too distant from the source distribution (e.g. test-time distribution is random), then the predicted labels tend to uniformly sample the label distribution: 
$\mathscr{f}_{\ell-1}(\theta_i; \mathbf{x}_j) \approx \{y: \frac{1}{|\mathcal{Y}|} \}$.
The low expected variance of the class probabilities $\sigma^2(\{y: \frac{1}{|\mathcal{Y}|} \}) \approx 0$ would suggest the expected accuracy w.r.t. perturbed inputs would also be $\frac{1}{|\mathcal{Y}|}$.
We assume the label set is finite and of small count, i.e. 
{\small
$\begin{cases}
|\mathcal{Y}| \nrightarrow \infty \\
\frac{1}{|\mathcal{Y}|} \nrightarrow 0
\end{cases}$
}.
For a small number of labels, we would expect an adversarially-perturbed input from a similar domain to return class probabilities skewed away from the ground-truth label 
{\small
$\max \texttt{dist}(\mathscr{f}_{\ell-1}(\theta_i; \mathbf{x}_{\textnormal{similar}}), \mathscr{f}_{\ell-1}(\theta_i; \mathbf{x}_{\textnormal{dissimilar}})) = \texttt{dist}(\{y: \rho_{\textnormal{similar}} \}, \{y: \rho_{\textnormal{dissimilar}} \})$
}
while also retaining non-zero variance $\sigma_{\textnormal{similar}}^2(\{y: \frac{1}{|\mathcal{Y}|} \}) > \sigma_{\textnormal{dissimilar}}^2(\{y: \frac{1}{|\mathcal{Y}|} \})$.
For a similar domain distribution that approximates or is near the source distribution in the input space with correspondingly low parameter distance (Proposition 2), 
as the parameters $\theta_i \approx \theta_j$, 
it follows that a (gradient-based) adversarial attack algorithm will be able to generate perturbations with respect to a close approximation of the parameters of the source distribution.
For a large number of labels where $|\mathcal{Y}| \rightarrow \infty$, this hypothesis may not hold, and a distant (dissimilar) distribution would attain an accuracy w.r.t. perturbed inputs of 0.
We formalize this result in Theorem 2.

\noindent
\mybox{\textbf{Evaluating Hypothesis 1} (Table \ref{tab:empirical.attack})}
We observe a significant gap between original accuracy and after-attack accuracy between different domain pairs.
The after-attack accuracy is worse than both the intra-attack accuracy and unperturbed accuracy, indicating a joint shift worsens accuracy w.r.t. perturbed inputs than each individual shift separately
(validating Hypothesis 1a).
Moreover, we observe a positive correlation between transfer loss and after-attack accuracy, and a negative correlation between SharedVocab and after-attack accuracy (albeit a low variance of distance).
Both indicate a joint shift manifesting similar domains lowers the accuracy further than that of dissimilar domains (validating Hypothesis 1b). 

Table \ref{tab:empirical.attack} also validates Proposition 1. 
The SharedVocab metric is high amongst all the domain pairs, and the variance between pairs is low.
The transfer loss across all domain pairs are low and below 0.1, also with minimal variance.
This empirically supports the notion that despite defined as being different domains of varying similarity, there is a common input space where all these distributions reside, and the distance between them is interpretatively low. 

\noindent\mybox{\textbf{Hypothesis 2}}
\textit{It is hypothesized 
the existence of a meta-learner that can compute high-accuracy parameters for inputs sampled from a joint-distributionally-shifted source distribution, 
where the perturbation sources are adversarial and domain-specific, 
and the input space is bounded; i.e. $\texttt{dist}(\mathbf{x}_i, \mathbf{x}_j) \mapsto \theta_i + \Delta \theta$. 
}

Proposition 2 indicates that a functional relationship may exist between $\texttt{dist}(X_i, X_j)$ and $\texttt{dist}(\theta_i, \theta_j)$.
Proposition 1 implies that, since $\mathcal{X}$ is a bounded space, $\texttt{dist}(X_i, X_j)$ and $\texttt{dist}(\theta_i, \theta_j)$ are bounded as well. 
Proposition 2 notes observations where the adaptation in parameters may not follow a linear pattern of increasing sparsity/density, increasing transferable/interfering subnetworks, etc; thus a non-linear adaptation function should be used. 

Suppose we construct a meta-learner $\mathscr{mf}$
that maps a change in input $X_j-X_i$ to a change in the parameters $\theta_j = \theta_i+\Delta \theta$, 
and the adapted parameters are used to perform inference in the base learner $\mathscr{f}$.

\begin{equation}
\small
\begin{split}
\Delta \theta & = \mathscr{mf}(\mathbf{x}_j) \\
\Delta \theta & \approx \mathscr{mf}(\texttt{dist}(\mathbf{x}_j, \mathscr{f}^{-1}(\theta_i))) \\
\Rightarrow \hat{y} & = \mathscr{f}(\theta_i+\Delta \theta ; \mathbf{x}_j)
\end{split}
\end{equation}

The meta learner may undergo different assumptions/constraints.
The meta learner may be provided task-specific / distribution-specific meta-data which encodes information parameter adaptation, usually used in few-shot learning implementations such as MAML \citep{finn2017modelagnostic}, Reptile \citep{nichol2018firstorder}, hypernetworks \citep{vonoswald2020continual}. 
In our setup, we do not presume any known information about the test-time distribution (i.e. no meta-data or headers).
The model is assumed to only have its own source distribution / dataset, and has access to no other domain data (including the attacker's domain). 
In many implementations, a meta-learner will accept $\mathbf{x}_j$ (and/or meta-data) as input arguments.
Meta-learners would need to compute a change in parameter distance $\Delta \theta \propto \texttt{dist}(\theta_i, \theta_j)$ based on $\texttt{dist}(\mathcal{X}_i, \mathcal{X}_j)$.
The underlying source distribution may not be hard-coded within $\mathscr{f}$ nor $\mathscr{mf}$, hence implicitly the source distribution would be approximated by inverting $\theta_i$ such that $\mathcal{X}_i \leftarrow \mathscr{f}^{-1}(\theta_i)$.

\noindent
\mybox{\textbf{Evaluating Hypothesis 2} (Tables \ref{tab:defenses} \& \ref{tab:defenses2})}
Table \ref{tab:defenses} validates Hypothesis 2. 
NLP models factor in semantic structure with respect to $\mathbf{P}_{i}$ through re-mapping unknown words to the $\texttt{<UNK>}$ token.
The ablation adversarial training defense outperforms the baseline adversarial training, indicating that constructing diverse, joint-distributionally-shifted perturbations sets in contrast to random perturbation sets yield marginal benefits.
Though both L2W and the ablation adversarial training make use of the perturbation sets, our results indicate, not only that a mapping can indeed be constructed between the changing shifts/perturbations and required parameter adaptation, but that computing an average/static parameter across varying perturbation sets is not optimal compared to enabling the base learner to adapt parameters according to the summation of probabilities of occurrence per sequence.
We additionally show in Table \ref{tab:defenses2} that our proposed defense is scalable across different model architectures and capacities.

\subsection{Theorems}
\label{theory3}

\noindent\textbf{Lemma 1. }
\textit{For an input variable $\mathbf{X}(\omega)$ that is sampled randomly, the output variable $X(\omega)$ from operations $\varepsilon$ applied to $\mathbf{X}(\omega)$ will also tend to be random.}

\noindent\textit{Proof. }
A random variable $\mathbf{X}$ is a mapping from $W$ to $\mathbb{R}$, that is $\mathbf{X}(\omega) \in \mathbb{R}$ for $\omega \in \mathbb{R}$.
$X(\omega) = \mathbf{X}(\omega) + \varepsilon$, thus $X$ is also a mapping $X: W \mapsto  \mathbb{R}$.
The measure for random variable $\mathbf{X}$ is defined by the cumulative distribution function $F(x) = \mathbf{P}(X \leq x)$.
For $x > 0$, 
$F_{X}(x) = \mathbf{P}(X \leq x) = \mathbf{P}(\mathbf{X}+\varepsilon \leq x) = \mathbf{P}(\mathbf{X} \leq x-\varepsilon) = F_{\mathbf{X}}(x-\varepsilon)$.
Thus $X(\omega)$ is also measurable and is a random variable defined on the sample space $W$.

\noindent\textbf{Lemma 2. }
\textit{Suppose a given model $\mathscr{f}(\theta; \mathbf{x}) = \theta \cdot \mathbf{x}$ and loss $\mathcal{L}(\theta; \mathbf{x},y)=\mathscr{f}(\mathbf{x})-y$.
Suppose we sample perturbations $(\mathbf{x_j} = \mathbf{x_i} + \xi), (y_i \rightarrow y_j) \sim \mathcal{X}, \mathcal{Y}$.
The change in loss between clean to perturbed input is $\frac{\partial \mathcal{L}}{\partial \theta} = \xi(\theta)+c$.}

\begin{proof}
\begin{equation}
\small
\begin{split}
\Delta\mathcal{L} & = \mathcal{L}(\theta; \mathbf{x_j}, y_i) - \mathcal{L}(\theta; \mathbf{x_i}, y) \\
& = [\mathscr{f}(\theta; \mathbf{x_j}, y_i) - \mathscr{f}(\theta; \mathbf{x_i}, y)] - [y_j-y_i] \\
&= \theta[\mathbf{x_j}-\mathbf{x_i}] - [y_j-y_i] \\
\frac{\partial^{2} \mathcal{L}}{\partial \theta^{2}} & = \mathbf{x_j}-\mathbf{x_i} = \xi \\
\frac{\partial \mathcal{L}}{\partial \theta} & = \xi(\theta)+c
\nonumber
\end{split}
\end{equation}
\end{proof}

\noindent\mybox{\textbf{Theorem 1}}
\textit{
Under joint distribution shift,
if the test-time distribution is too distant from the source distribution, then the predicted labels tend to uniformly sample the label distribution: 
$\mathscr{f}_{\ell-1}(\theta_i; \mathbf{x_j}) \approx \{y: \frac{1}{|\mathcal{Y}|} \}$.
}

\noindent\textbf{Proof sketch of Theorem 1. }
Suppose we sample perturbations $(\mathbf{x_j} = \mathbf{x_i} + \xi), (y_i \rightarrow y_j) \sim \mathcal{X}, \mathcal{Y}$.

\begin{equation}
\small
\begin{split}
\mathbf{x_j} & = \mathbf{x_i} + \xi \\
\mathcal{L}(\mathbf{x_j}, y) & = \mathcal{L}(\mathbf{x_i}, y) + \mathcal{L}(\xi, y) \\
\frac{\partial \mathcal{L}(\mathbf{x_j}, y)}{\partial \theta} & = \frac{\partial \mathcal{L}(\mathbf{x_i}, y)}{\partial \theta} + \frac{\partial \mathcal{L}(\xi, y)}{\partial \theta} \\
\Rightarrow 
\theta_{t} & := \theta_{t-1} 
- \sum_{\mathbf{x_i}, y}^{\mathcal{X}, \mathcal{Y}} \frac{\partial \mathcal{L}(\mathbf{x_i}, y)}{\partial \theta}
- \sum_{\xi, y}^{\mathcal{X}, \mathcal{Y}} \frac{\partial \mathcal{L}(\xi, y)}{\partial \theta}
\nonumber
\end{split}
\end{equation}

This decomposition implies
$\frac{\partial \mathcal{L}(\mathbf{x_i}, y)}{\partial \theta}$ updates part of $\theta$ w.r.t. $\mathbf{x_i}$, which we denote as $\theta \odot m_{\mathbf{x_i}}$,
and 
$\frac{\partial \mathcal{L}(\xi, y)}{\partial \theta}$ updates part of $\theta$ w.r.t. $\xi$, which we denote as $\theta \odot m_{\xi}$,
where $m_{\mathbf{x_i}}, m_{\xi} \in \{ 0,1 \}^{|\theta|}$ are masks of $\theta \equiv \theta \odot (m_{\mathbf{x_i}} + m_{\xi})$.
Given the distances (squared Euclidean norm) between the shifted inputs and outputs $\mathbf{x_i} \rightarrow \mathbf{x_j}$ and $y_i \rightarrow y_{j}$, we can enumerate the following 4 cases.
Case (1) is approximates minimal or negligible distribution shift, and is not evaluated.
As the scope of (joint) distribution shift specifies a change in the input, a lack of shift in the input distribution invalidates consideration of Case (3). 

{\small
\begin{numcases}{}
\begin{split}
\small
& || \mathbf{x_j} - \mathbf{x_i} ||_2^2 \approx 0 \phantom{=},\phantom{=} || y_j - y_i ||_2^2 \approx 0
\phantom{==} \textnormal{(Case 1)}
\end{split}
\nonumber
\\
\begin{split}
\small
& || \mathbf{x_j} - \mathbf{x_i} ||_2^2 > 0 \phantom{=},\phantom{=} || y_j - y_i ||_2^2 \approx 0
\phantom{==} \textnormal{(Case 2)}
\end{split}
\nonumber
\\
\begin{split}
\small
& || \mathbf{x_j} - \mathbf{x_i} ||_2^2 \approx 0 \phantom{=},\phantom{=} || y_j - y_i ||_2^2 > 0
\phantom{==} \textnormal{(Case 3)}
\end{split}
\nonumber
\\
\begin{split}
\small
& || \mathbf{x_j} - \mathbf{x_i} ||_2^2 > 0 \phantom{=},\phantom{=} || y_j - y_i ||_2^2 > 0
\phantom{==} \textnormal{(Case 4)}
\end{split}
\nonumber
\end{numcases}
}

We denote a random distribution $\texttt{Rand}: s \sim \mathcal{U}(S)$ s.t. $\mathbb{P}(s)=\frac{1}{|S|}$, 
where an observation $s$ is uniformly sampled from (discrete) set $S$.
For a set of perturbations per sequence $\xi = \sum_{\lambda}^{\Lambda} \varepsilon_{\lambda}$, where $\Lambda$ are different sources/variations of shift, 
if $\varepsilon_{\lambda} \rightarrow d_{\textnormal{max}}$ and/or
$\sum_{\lambda}^{\Lambda \rightarrow \infty} \varepsilon_{\lambda} \rightarrow d_{\textnormal{max}}$ such that $\xi \sim \texttt{Rand}$,
then 
$\frac{\partial \mathcal{L}(\theta; \xi, y)}{\partial \theta} \sim \texttt{Rand}$
and $\mathscr{f}(\theta; \mathbf{x_j}) - \mathscr{f}(\theta; \mathbf{x_i}) \approx \mathscr{f}(\theta; \xi) \sim \texttt{Rand}$ (by Lemma 1 and 2). 

\noindent
Hence, for each case of $\frac{\partial \mathcal{L}(\theta; \xi, y)}{\partial \theta}$:

if \phantom{} $\frac{\partial \mathcal{L}(\theta; \xi, y)}{\partial \theta} \neq 0$, given $\theta = \theta \odot (m_{\mathbf{x_i}} + m_{\xi})$, then $\mathscr{f}(\theta; \xi) \approx \mathscr{f}(\theta+ m_{\xi}; \xi) \sim \texttt{Rand}$;

if \phantom{} $\frac{\partial \mathcal{L}(\theta; \xi, y)}{\partial \theta} = 0$, given $m_{\mathbf{x_i}} = 1^{|\theta|}, m_{\xi} = 0^{|\theta|}$, then $\mathscr{f}(\theta; \xi) \approx \mathscr{f}(\xi; \theta+ m_{\mathbf{x_i}}) \sim \texttt{Rand}$.

In both cases, the predicted value of $\mathscr{f}$ will be sampled randomly.
Given it randomly samples from the label space $\mathcal{Y}$ for distributionally-shifted input:output Cases (2) and (4), 
it follows that,
under the presence of distant (joint) distribution shift at test-time,
a prediction $y \sim \mathcal{U}(\mathcal{Y})$ s.t.
$\mathscr{f}_{\ell-1}(y; \{\theta_i; \mathbf{x_j}\}) = \frac{1}{|\mathcal{Y}|}$.
For the $(\ell-1)$th layer that computes class probabilities, this results in $\mathscr{f}_{\ell-1}(\theta_i; \mathbf{x_j}) \approx \{y: \frac{1}{|\mathcal{Y}|} \}$.

\qed

\noindent\mybox{\textbf{Theorem 2}}
\textit{
A joint shift with a similar domain and adversarial perturbations ($0 \leq \xi \leq \infty$) will return a lower accuracy when the number of classes is finite and bounded w.r.t. the class probability w.r.t. $\xi$ at $|\mathcal{Y}| < \frac{1}{\rho(\xi)}$; if $|\mathcal{Y}| \rightarrow \infty$, then a joint shift with a dissimilar domain and adversarial perturbations ($\xi \rightarrow \infty$) will return a lower accuracy w.r.t. perturbed inputs.
}

\noindent\textbf{Proof sketch of Theorem 2. }
We simplify $\mathscr{f}_{\ell-1}(\theta; \mathbf{x}) = \{ y : \rho\}$ as $\mathscr{f}_{\ell-1}(y; \{\theta; \mathbf{x}\}) = \rho$ to compute the class probability of label $y$.
Perturbations $\xi$ lower the class probabilities to $\rho(\xi)$ below the clean $\rho$:
$\mathscr{f}_{\ell-1}(y; \{\theta; \mathbf{x}+\xi \}) = \rho(\xi) < \rho$.
We simplify our analysis, such that when $\rho(\xi) < \rho$, then $\mathscr{f}$ predicts $\neg y$ where $\neg y \neq y$, i.e. a lower $\rho(\xi)$ returns a lower accuracy w.r.t. perturbed inputs.

We consider 3 cases to evaluate how to maximize $\Delta \rho = \mathscr{f}_{\ell-1}(y; \{\theta; \mathbf{x}+\xi \}) - \mathscr{f}_{\ell-1}(y; \{\theta; \mathbf{x} \})$:

{\small
\begin{numcases}{}
\begin{split}
\small
& 
\Delta \rho_{\xi = 0}
\approx 0
\end{split}
\nonumber
\\
\begin{split}
\small
& 
\Delta \rho_{0 \leq \xi \leq \infty}
= \rho(\xi) - \rho < 0
\end{split}
\nonumber
\\
\begin{split}
\small
& 
\Delta \rho_{\xi \rightarrow \infty}
\approx \frac{1}{|\mathcal{Y}|} - \rho < 0
\phantom{==} \textnormal{\textit{(Thm 1)}}
\end{split}
\nonumber
\end{numcases}
}

In order for 
perturbations w.r.t. a similar domain to reduce the accuracy lower than that w.r.t. a dissimilar domain
$|\Delta \rho_{0 \leq \xi \leq \infty}| > |\Delta \rho_{\xi \rightarrow \infty}|$,
the number of classes $|\mathcal{Y}|$ needs to be greater than $\frac{1}{\rho(\xi)}$ for a given $\xi$:

\begin{equation}
\small
\begin{split}
\Delta \rho_{0 \leq \xi \leq \infty} & < \Delta \rho_{\xi \rightarrow \infty} \\
\rho(\xi) & < \frac{1}{|\mathcal{Y}|} \\
|\mathcal{Y}| & < \frac{1}{\rho(\xi)}
\end{split}
\nonumber
\end{equation}

Given the bounds of the class probability w.r.t. $y$, 
we can conclude that a joint shift with a similar domain and adversarial perturbations ($0 \leq \xi \leq \infty$) will return a lower accuracy when the number of classes is finite and bounded at $|\mathcal{Y}| < \frac{1}{\rho(\xi)}$; if $|\mathcal{Y}| \rightarrow \infty$, then a joint shift with a dissimilar domain and adversarial perturbations ($\xi \rightarrow \infty$) will return a lower accuracy w.r.t. perturbed inputs.

{\small
\begin{align}
0 \leq & \rho(\xi) < \frac{1}{|\mathcal{Y}|} \Bigg|_{|\mathcal{Y}| < \frac{1}{\rho(\xi)}} \notag\\
\frac{1}{|\mathcal{Y}|} \Bigg|_{|\mathcal{Y}| \rightarrow \infty} \leq & \rho(\xi) < \frac{1}{|\mathcal{Y}|} \Bigg|_{|\mathcal{Y}| < \frac{1}{\rho(\xi)}}
\end{align}}

\qed

\end{document}